\documentclass{article}

\usepackage[preprint]{neurips_data_2024}

\usepackage{natbib}
\usepackage[utf8]{inputenc} 
\usepackage[T1]{fontenc}    
\usepackage{hyperref}       
\usepackage{url}            
\usepackage{booktabs}       
\usepackage{amsfonts}       
\usepackage{nicefrac}       
\usepackage{microtype}      

\usepackage{graphicx}
\usepackage{courier}
\usepackage{amsmath}
\usepackage{amssymb}
\usepackage{bbm}
\usepackage{adjustbox}
\usepackage{pifont} 
\usepackage{xcolor} 

\usepackage{multirow}
\usepackage{multicol}
\usepackage{changepage,threeparttable} 
\usepackage{makecell}
\usepackage{colortbl}
\usepackage{xspace}
\usepackage{enumitem}
\usepackage{subfigure}
\usepackage{wrapfig}

\usepackage{utfsym}
\newcommand{\cmark}{\usym{2714}}
\newcommand{\xmark}{\usym{2717}}

\title{MedOdyssey: A Medical Domain Benchmark for \\ Long Context Evaluation Up to 200K Tokens}

\author{Yongqi Fan\textsuperscript{\rm $\diamondsuit$\thanks{~~Co-first authors.}}, 
 Hongli Sun\textsuperscript{\rm $\diamondsuit$\footnotemark[1]}, 
 Kui Xue\textsuperscript{\rm $\spadesuit$},
 Xiaofan Zhang\textsuperscript{\rm $\heartsuit\spadesuit$},
 Shaoting Zhang\textsuperscript{\rm $\spadesuit$},
 Tong Ruan\textsuperscript{\rm $\diamondsuit$}\thanks{~~Corresponding author.} \\
\textsuperscript{\rm $\diamondsuit$}School of Information Science and Engineering, East China University \\ of Science and Technology, Shanghai, China \\
\textsuperscript{\rm $\spadesuit$}Intelligent Healthcare, Shanghai Artificial Intelligence Laboratory, Shanghai, China \\
\textsuperscript{\rm $\heartsuit$}School of Electronic Information and Electrical Engineering,\\ Shanghai Jiao Tong University, Shanghai, China \\
\texttt{\{johnnyfans,shl\}@mail.ecust.edu.cn} \\
\texttt{\{xuekui,zhangshaoting\}@pjlab.org.cn} \\
\texttt{xiaofan.zhang@sjtu.edu.cn}, \texttt{ruantong@ecust.edu.cn} \\
}

\begin{document}

\maketitle

\begin{abstract}
Numerous advanced Large Language Models (LLMs) now support context lengths up to 128K, and some extend to 200K. Some benchmarks in the generic domain have also followed up on evaluating long-context capabilities. In the medical domain, tasks are distinctive due to the unique contexts and need for domain expertise, necessitating further evaluation. However, despite the frequent presence of long texts in medical scenarios, evaluation benchmarks of long-context capabilities for LLMs in this field are still rare. In this paper, we propose MedOdyssey, the first medical long-context benchmark with seven length levels ranging from 4K to 200K tokens. MedOdyssey consists of two primary components: the medical-context ``needles in a haystack'' task and a series of tasks specific to medical applications, together comprising 10 datasets. The first component includes challenges such as counter-intuitive reasoning and novel (unknown) facts injection to mitigate knowledge leakage and data contamination of LLMs. The second component confronts the challenge of requiring professional medical expertise. Especially, we design the ``Maximum Identical Context'' principle to improve fairness by guaranteeing that different LLMs observe as many identical contexts as possible. Our experiment evaluates advanced proprietary and open-source LLMs tailored for processing long contexts and presents detailed performance analyses. This highlights that LLMs still face challenges and need for further research in this area. Our code and data are released in the repository: \url{https://github.com/JOHNNY-fans/MedOdyssey.}
\end{abstract}

\section{Introduction}
\label{sec:intro}
Long-Context Large Language Models (LLMs)~\citep{GPT-4, claude-model-card, ai2024yi} have become a mainstream research topic. Using the advanced Transformer architecture, position embedding, and other techniques~\citep{huang2023advancing, peng2023yarn, jin2024llm, ding2024longrope}, LLMs' context length (context window) is extended, and long-context prompts frequently encountered in practical scenarios can be supported to handle, such as books, lengthy chat history or documents retrieved from website. The LLMs currently available on the market generally support context lengths of 8k tokens. Advanced models have extended this capability to 128k tokens, with some even reaching 200k tokens or more. Researchers have swiftly responded by conducting evaluations of LLMs in long contexts, proposing numerous works in the generic domain to assess their performance. These include the classic and funny needle-in-a-haystack experimental projects~\citep{gkamradt2024llmtest, song2024countingstars} and several benchmarks~\citep{an2023eval, yuan2024lv, zhang2024infty} for evaluating and analyzing the long-context performance of LLMs.

In the medical domain, evaluating the medical capabilities of LLMs is often conducted independently due to the unique context and the need for professional knowledge~\citep{tang2023evaluating, jin2021disease, zhu2023promptcblue, singhal2023large}. However, the long-context evaluations in this field~\cite {saab2024capabilities} are relatively infrequent and lack of medical-context ``needles in a haystack'' experiment. Actually, there are some more difficult long-context scenarios that exist for medical practices, e.g., biomedical terminology normalization and electronic health record (EHR) analysis~\citep{sarker2018data, shickel2017deep}. There is a noticeable lack of benchmarks involving a package of basic and various long-context evaluation tasks.

\begin{figure}[t]
    \centering
    \includegraphics[width=\linewidth]{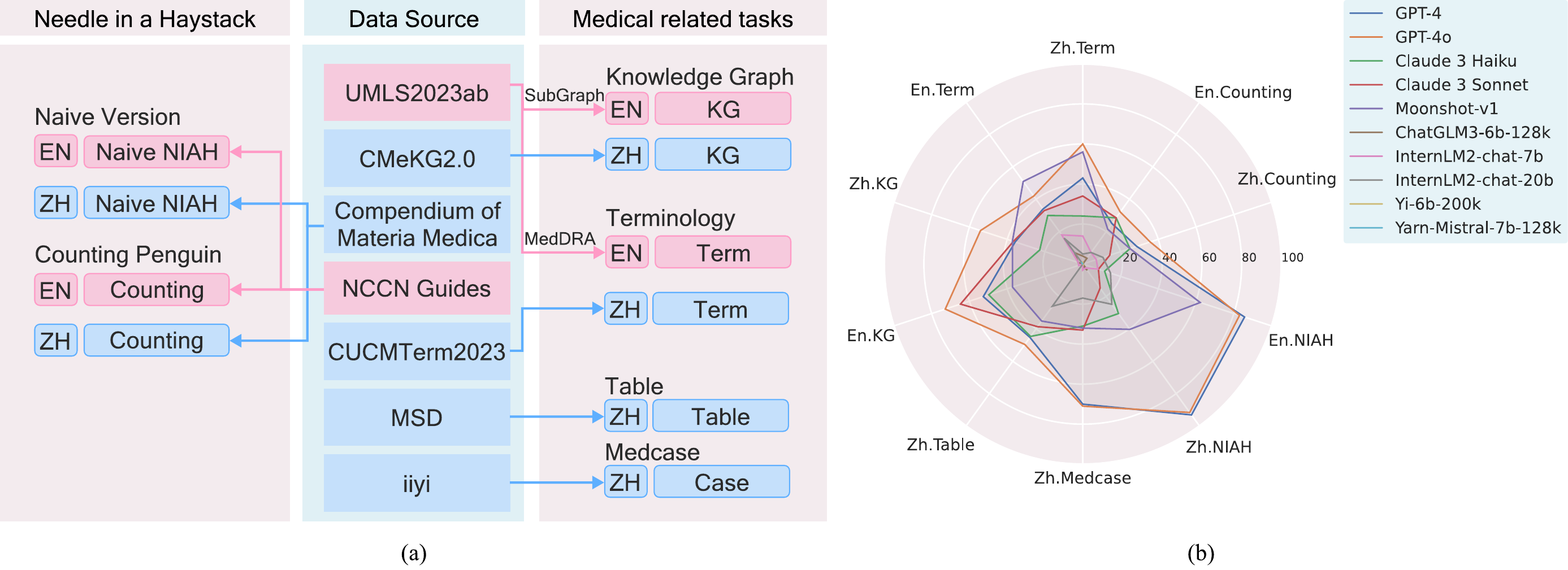}
    \caption{(a). The overall architecture of the MedOdyssey. (b).Radar chart of the overall performance of long-context LLMs on MedOdyssey.}
    \label{fig:datasets_and_result_radar}
\end{figure}

In this paper, we propose MedOdyssey, the first medical-domain long-context evaluation benchmark for LLMs. MedOdyssey is comprised of two primary components: the medical-context needles in a haystack (NIAH) tasks and a series of medical-related tasks, containing 10 complex datasets and involving several medical domain professional corpora, e.g., medical books and guides, medical cases with electronic health records, medical knowledge graphs, medical terminology database and medical tables. Based on these corpora, we construct several evaluation tasks, as shown in part (a) of Figure~\ref{fig:datasets_and_result_radar}. Additionally, apart from the naive implementation, we introduced the latest Counting Stars~\citep{song2024countingstars} to enhance the reliability of the ``needle in a haystack'' component. To ensure fairness, we propose a new ``maximum identical context'' principle to address the issue of varying contexts resulting from direct middle truncation~\citep{zhang2024infty, yuan2024lv}. We also prevent data contamination and data leakage during evaluation by incorporating counter-intuitive reasoning problems and novel (unknown) facts questions.

We evaluate the performance of advanced LLMs remarkably supporting long-context prompts, including both proprietary and open-source models. The overall performance is shown in part (b) of Figure~\ref{fig:datasets_and_result_radar} using a radar chart. Our experimental results demonstrate that the performance of LLMs in the medical long contexts is actually still lacking.
Specifically, even the newest GPT-4o only performs well in the naive NIAH experiment, and is not a hexagonal warrior. Moreover, we perform a comprehensive analysis to provide insights and direction. We encourage further research by the NLP community to jointly address the more realistic settings presented in this benchmark.
\section{Related Work}
\textbf{Long-Context LLMs.}
The challenge of supporting long-context prompts for LLMs has been a focal research topic, leading to various innovative approaches. Numerous position embedding methods and efficient transformer architectures have been instrumental in extending the maximum context length of LLMs. Notable examples include Roformer~\citep{SU2024127063}, ALiBi~\citep{press2022train}, Longformer~\citep{beltagy2020longformer}, Reformer~\citep{kitaev2020reformer}, and LM-Infinite~\citep{han2023lm}. Recent advancements in long-context LLMs have garnered significant interest, with many models now supporting extended context lengths of up to 128K tokens or more. For instance, GPT-4~\citep{GPT-4}, Moonshot~\citep{moonshotAI}, Yarn-Mistral~\citep{peng2023yarn}, and ChatGLM3~\citep{ChatGLM3} can handle up to 128K tokens. Furthermore, models such as Claude 3~\citep{claude-model-card} and Yi~\citep{ai2024yi} support context lengths up to 200K tokens.

\textbf{Generic-domain Long-Context Evaluation for LLMs.}
Some research focuses on the capability of LLMs to process long contexts, proposing various datasets and benchmarks. For example, ZeroSCROLLS~\citep{shaham-etal-2023-zeroscrolls} evaluates state-of-the-art LLMs through document summarization, question answering, and aggregation tasks. L-Eval~\citep{an2023eval} relabeled some public datasets and proposed additional evaluation metrics. LongBench~\citep{bai2023longbench} constructed a comprehensive benchmark with six major task categories, each varying in length, language, and domain. BAMBOO~\citep{dong-etal-2024-bamboo-comprehensive} addresses the issue of data contamination by using only data released in 2023. However, most of these studies do not include evaluations in the medical domain.

\textbf{Medical-domain Evaluation Benchmark for LLMs.}
LLMs are increasingly used in medical fields, where specialized and dense information requires different evaluation methods than general domains. Some studies focus on specific capabilities within the medical field. For instance, \citet{tang2023evaluating} assess LLMs with zero-shot medical evidence summarization, and \citet{doi:10.1056/AIoa2300151} evaluate LLMs in specific medical areas. Primary data sources often include existing exams or benchmarks. \citet{jin2021disease} created the MedQA dataset from medical board exams. PromptCBLUE~\citep{zhu2023promptcblue} evaluates Chinese LLMs using a baseline based on CBLUE, avoiding corpus leakage. \citet{NEURIPS2023_a48ad12d} uses questions from the Chinese National Medical Licensing Examination, while MultiMedQA~\citep{singhal2023large} combines six medical QA datasets from online searches. However, there is a lack of evaluation benchmarks for the medical long context.

\begin{figure}[ht]
    \centering
    \includegraphics[width=\linewidth]{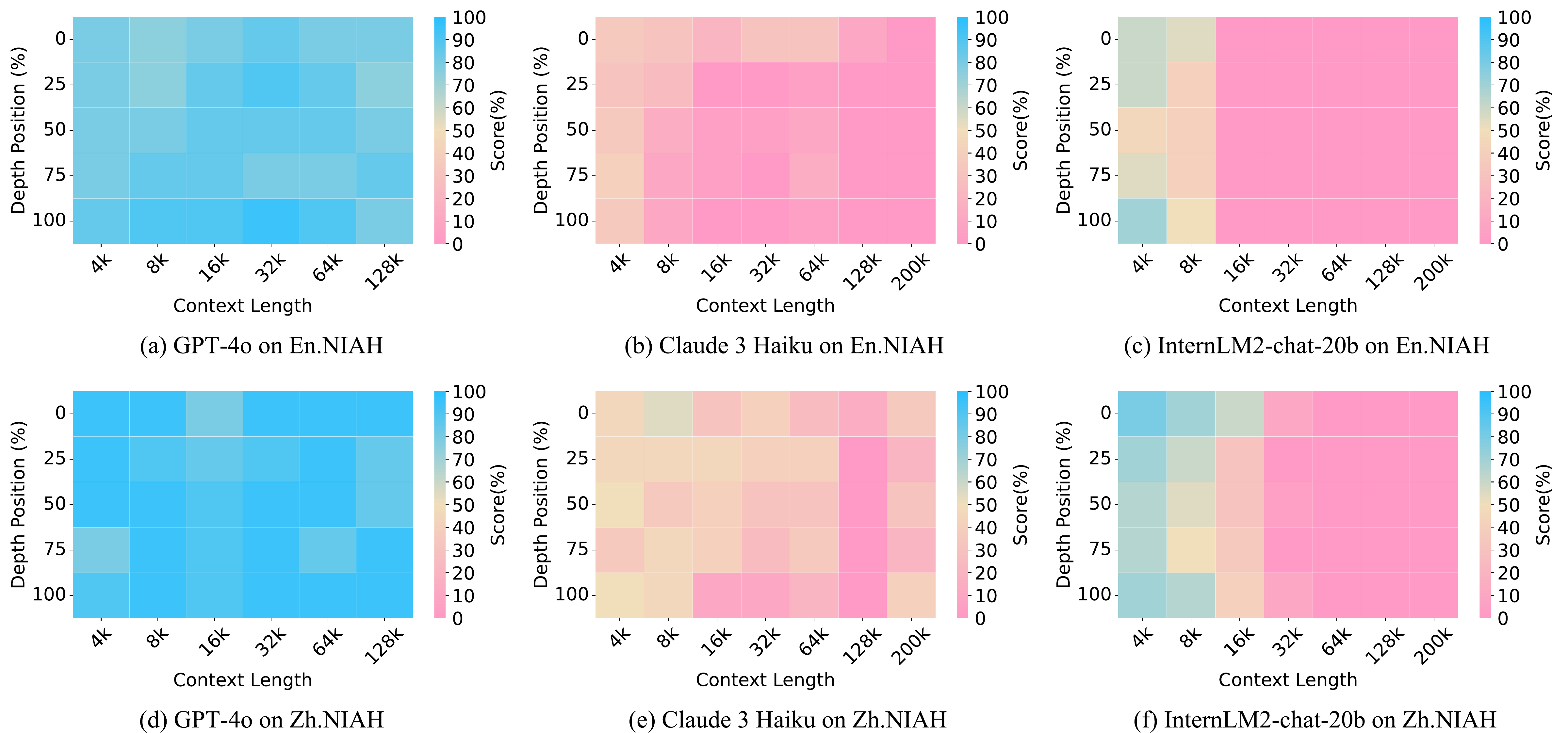}
    \caption{Heatmaps of GPT-4o, Claude 3 Haiku and InternLM2-chat-20b on NIAH task.}
    \label{fig:niah_heatmap}
\end{figure}
\section{The MedOdyssey Benchmark}
\label{sec:bench}
\subsection{Dataset Collection}
Due to copyright and privacy protection concerns, collecting diverse and valuable corpora is challenging. We opted not to use simulation, self-building, or distillation techniques due to the need for medical expertise. Consequently, we dedicated significant effort to finding academic open-source, formal application pathways, and copyright-free medical data and knowledge. As shown in part (a) of Figure~\ref{fig:datasets_and_result_radar}, for the ``needles in a haystack'' part, we have collected 30 volumes of Chinese medical books ``Compendium of Materia Medica'' from an open-source repository\footnote{\url{https://github.com/lab99x/tcmoc/tree/master}}, and three English clinical guides\footnote{\url{https://www.nccn.org/guidelines/}} in PDF format were converted to meet long text requirements. And there are four knowledge bases involved in medical-related tasks. We converted and organized the ``Chinese Common Clinical Medical Terminology 2023 Edition'' (CUCMTerm2023) from PDF format to obtain four types of standard terms: disease diagnosis, clinical examination, procedure operation, and symptom. We used MedDRA terms from the UMLS2023ab version~\citep{bodenreider2004unified}\footnote{\url{https://www.nlm.nih.gov/research/umls/}} as the foundational terminology bases. Additionally, we used CMeKG2.0\footnote{\url{http://cmekg.pcl.ac.cn/}} and extracted MedDRA subgraphs from the UMLS2023ab version as the basic knowledge graphs. We also obtained 500 medical cases with EHR information from an open-source medical forum iiyi\footnote{\url{https://bingli.iiyi.com/}}, and 100 medical tables from an open-source medical website MSD\footnote{\url{https://www.msdmanuals.cn/professional/pages-with-widgets/tables?mode=list}}, via spider techniques.

\begin{table}[ht]
    \centering \small
    \caption{Dataset statistics and descriptions. The columns indicate the annotation method (auto-generated or human), the number of examples, average text length (input/output), use of the construction strategy from Section~\ref{sec: construction}, and the evaluation metrics.} 
    \resizebox{1.0\linewidth}{!}{
    \begin{tabular}{l|ccc|ccc|c}
        \toprule
        \textbf{Task} & \textbf{Annotation} & \textbf{\# Examples} & \textbf{Avg. Len} & \textbf{MIC} & \textbf{NFI} & \textbf{CIR} & \textbf{Eval Metrics} \\
        \midrule
        En.NIAH & Auto \& Human & 20$\times$7$\times$5 & 179.2k/32 & $\cmark$ & $\cmark$ & $\xmark$ & Acc. \\
        Zh.NIAH & Auto \& Human & 20$\times$7$\times$5 & 45.6k/10.2 & $\cmark$ & $\cmark$ & $\xmark$ & Acc. \\
        En.Counting & Auto & 4$\times$7 & 179.0k/13.6 & $\cmark$ & $\xmark$ & $\cmark$ & Acc. \\
        Zh.Counting & Auto & 4$\times$7 & 45.6k/12.3 & $\cmark$ & $\xmark$ & $\cmark$ & Acc. \\
        \midrule
        En.KG & Auto \& Human & 100 & 186.4k/68.8 & $\cmark$ & $\xmark$ & $\cmark$ & P., R., F1. \\
        Zh.KG & Auto \& Human & 100 & 42.5k/2.0 & $\cmark$ & $\xmark$ & $\cmark$ & P., R., F1. \\
        En.Term & Auto & 100 & 183.1k/11.7 & $\cmark$ & $\xmark$ & $\xmark$ & Acc. \\
        Zh.Term & Auto & 100 & 32.6k/7.0 & $\cmark$ & $\xmark$ & $\xmark$ & Acc. \\
        Zh.Case & Auto \& Human & 100 & 47.7k/1.3 & $\cmark$ & $\xmark$ & $\xmark$ & Acc. \\
        Zh.Table & Auto \& Human & 100 & 53.6k/1.4 & $\cmark$ & $\xmark$ & $\xmark$ & P., R., F1. \\
        \bottomrule
    \end{tabular}}
    \label{tab:data-statistics}
\end{table} 
\subsection{Shard Task in MedOdyssey}
Based on these collected corpora, we define a total of ten shared tasks in two broad categories needles in a haystack and medical-related tasks, as shown in Figure~\ref{fig:datasets_and_result_radar}. 

\textbf{NIAH.} The naive need in a haystack, inserting a fragment of unrelated knowledge (the needle) within a lengthy text (the haystack) and then prompting the LLM to answer questions about the it.

\textbf{Counting.} This is a more challenging variation of the NIAH task. Within the context of a virtual story, dispersed counting fragments are embedded throughout a lengthy text. The LLM is then prompted to identify and output the sequence of these counting fragments.

\textbf{Term Norm.} The medical terminology normalization task, requires LLMs to identify the corresponding standard term for a medical phrase from a large standard terminology database.

\textbf{KG QA:} The LLM is prompted to answer questions derived from a medical knowledge graph presented in triplet form, concentrating on the relationships of entities and relationships.

\textbf{Table QA:} This task involves the LLM responding to questions based on medical tables that are formatted in Markdown.

\textbf{Case QA:} Here, the LLM addresses questions related to provided medical cases, which include details of patient EHR information and the treatment processes.

We use some Chinese books and English guides as the haystack in NIAH and Counting tasks. Additionally, all QA tasks are based on closed-ended, text-based questions. Figure~\ref{fig:NIAH_example} to Figure~\ref{fig:table_case_example} shows the examples of input and output.
\subsection{Construction and Annotation}
\label{sec: construction}
Our benchmark is primarily designed to evaluate the long-context capabilities of LLMs within medical texts. By examining the context windows supported by advanced LLMs, we have selected seven token lengths: 4k, 8k, 16k, 32k, 64k, 128k, and 200k. Throughout the dataset construction, we employed both automated and manual collaboration, implementing various strategies to support fairness and professionalism. These measures also aimed to minimize the impact of data contamination and leakage. The next step is to present each point in detail.

\begin{figure}[ht]
    \centering
    \includegraphics[width=\linewidth,height=0.3\linewidth]{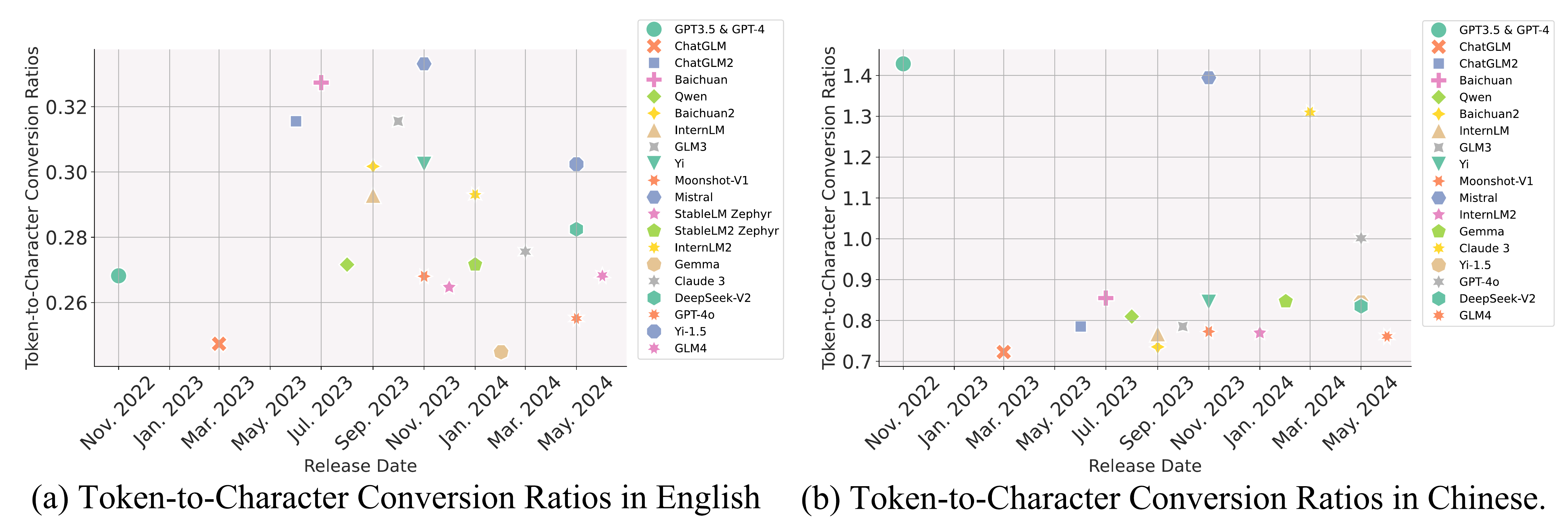}
    \caption{Trends in token-to-character conversion rates for advanced LLMs over time.}
    \label{fig:all_token_conversion}
\end{figure}
\textbf{Maximum Identical Context (MIC).}
It is worth noting that the current work aims to reach the maximum number of tokens for a given model, employing intermediate truncation when performing long-context evaluation. In practice, this strategy results in different models receiving different contextual texts, which ultimately lacks fairness.

In our work, we present the ``Maximum Identical Context'' principle and convert a fixed number of tokens to a fixed range. By analyzing the token-to-character conversion ratios of advanced LLMs in Table~\ref{tab:token_scale_ratio}, we select a fixed conversion ratio for both Chinese and English to ensure that all LLMs can see the same context while accepting the maximum number of tokens. Formally, our goal is to optimize the formula~\ref{equ: max_context_length} for each sample to obtain the maximum text length $L'$ corresponding to a certain number of tokens $N$, where $C$ is the predefined token length list and $\gamma$ is the specific maximum token-to-character conversion ratio analyzed from Figure~\ref{fig:all_token_conversion} and Table~\ref{tab:token_scale_ratio}. In practice, all our dataset builds adopt this principle to get the maximum identical context shared across LLMs. 

We acknowledge that adding a new LLM during our competition could impact the token-to-character conversion ratio and the dataset. Nonetheless, we remain committed to this approach and have identified effective measures through risk analysis to address these challenges. As shown in Figure~\ref{fig:all_token_conversion}, a clear trend is that the token-to-character conversion ratio of advanced LLMs is decreasing, which will keep our benchmark robust. Meanwhile, we tend to integrate \textbf{MedOdyssey} into periodic evaluation platforms, adjusting it by periodically adapting to new token-to-character conversion ratios, replacing old questions with new ones, and using code automation to complete the build. This approach will help further ensure fairness and prevent data leakage.

\begin{equation}
\min_{N \in C} \left( \frac{N}{\gamma} - L' \right), \quad L' \leq \frac{N}{\gamma}, \quad \text{where } C = \{4k, 8k, \ldots, 200k\}
\label{equ: max_context_length}
\end{equation}

\textbf{Novel Facts Injection (NFI).}
To prevent data leakage and contamination, i.e., to ensure that LLMs have not been trained on question-related data, we employ a novel fact injection method in the naive needle-in-a-haystack task. Specifically, we manually and meticulously crafted needles and their corresponding questions for the needle-in-a-haystack task, including ten non-medical questions and ten medical questions. These twenty questions are based on the latest information, with the general portion drawn from the newest plot and setting of the ``Honkai: Star Rail'' game, and the medical portion sourced from the latest literature in The Lancet and some real doctor-patient dialogues. Meanwhile, in this task, we measure the effect of five different depths at which the needle is located and seven different lengths of the haystack, achieved through automated code execution. Eventually, we get the datasets \textbf{En.NIAH} and \textbf{Zh.NIAH}.

\textbf{Counter-intuitive Reasoning (CIR).}
Acquiring systematic medical knowledge, such as knowledge graphs, is challenging due to the slow accumulation of medical information. To address the difficulty in ensuring that the model hasn't been trained on this type of knowledge, we introduced counter-intuitive designs to test the LLM's reasoning with long contexts. For example, in the KG task, we ask the model to find all the triples that can answer a question instead of directly providing an answer. We randomized some questions involving three cases from the graph: head-entity to tail-entity, head-entity to relationship, and relationship to tail-entity, and generated questions using pre-constructed templates. These questions were then rewritten with guaranteed semantics using the state-of-the-art LLM GPT-4o~\citep{GPT-4o}. For a given sample, we identify all relevant triples as the correct answer based on all input triples, resulting in the dataset \textbf{En.KG} and \textbf{Zh.KG}.

Similarly in the counting task, we designed a counter-intuitive story setting, i.e., we have a little star count penguins, where the LLM must retain the memory of the task goal regardless of the context length. Additionally, For the ``Counting Penguin'' task, four different difficulty types were designed, including counting a penguin repeatedly, counting penguins incrementally, counting penguins disorderly, and counting penguins with corrections. As in the original project, we use the correct counting order as the answer, and we get the dataset \textbf{En.Counting} and \textbf{Zh.Counting}.

\textbf{Others.}
During the development of our terminology normalization task, we adopt a reputable dataset from the second shard task of the SMM4H-17\footnote{\url{https://data.mendeley.com/datasets/rxwfb3tysd/1}}~\citep{sarker2018data}. challenge for the English language and dataset the \textbf{En.Term} dataset. For the Chinese language, we have independently constructed a dataset \textbf{Zh.Term} for terminology normalization based on the synonyms and previously utilized phrases in CUCMTerm2023 corpus, which includes the same four term categories present in our established standard terminology database.

For both the medical table QA dataset \textbf{Zh.Table} and medical case QA dataset \textbf{Zh.Case}, we use a manual querying strategy by randomly selecting a medical table or case and formulating questions based on the relevant information it contains. For example, when working with a medical table, we ask questions related to the specific medical knowledge presented in the table. In the context of medical cases, our questions cover aspects such as the patient's chief complaint, symptoms, result of imaging studies, findings of complete checkup.

\textbf{Dataset Statistics and Descriptions.}
We present the dataset statistics and the general overview in Table~\ref{tab:data-statistics}, where ``MIC'' is the short of \textbf{M}aximum \textbf{I}dentical \textbf{C}ontext, ``NFI'' is the short of \textbf{N}ovel \textbf{F}acts \textbf{I}njection, and ``CIR'' is the short of \textbf{C}ounter-\textbf{i}ntuitive \textbf{R}easoning.
\section{Experiments}
\label{sec:ex}
\subsection{Baseline Models}
We researched current state-of-the-art long-context LLMs and presented the performance of two kinds of baseline LLMs in MedOdyssey. For closed-source commercial LLMs, we call the official APIs to get the responses for each task. We also deployed open-source models for inference on our own. The LLMs and versions we selected are as follows:

\textbf{GPT-4}~\citep{GPT-4}: Released in March 2023, GPT-4 is a state-of-the-art language model developed by OpenAI. It supports a context window length of 8,192 tokens, which was extended to 128k in the November 2023 update. (gpt-4-turbo-2024-04-09)

\textbf{GPT-4o}~\citep{GPT-4o}: An optimized variant of GPT-4, GPT-4o was introduced in May 2024, has a 128k context window, and has a knowledge cut-off date of October 2023. (gpt-4o-2024-05-13)

\textbf{Claude 3}~\citep{claude-model-card}: Launched by Anthropic in March 2024, the family includes three models in ascending order of capability: Haiku, Sonnet, and Opus, allowing users to select. The three models offer a 200k context window upon launch. (claude-3-haiku-20240307 and claude-3-sonnet-20240229)

\textbf{Moonshot-v1}~\citep{moonshotAI}: Released in 2023 by Moonshot AI, it emphasizes scalability and supports a context window of 128k tokens for generating very long texts. (moonshot-v1-128k)

\textbf{ChatGLM3-6b-128k}~\citep{ChatGLM3}: Developed by ZHIPU·AI in 2024, it builds based on ChatGLM3-6B and better handles long contexts up to 128K tokens.

\textbf{InternLM2}~\citep{cai2024internlm2}: An open-source LLM is introduced in 2024 by Shanghai AI Lab, including 7b and 20b sizes. It initially trained on 4k tokens before advancing to 32k tokens in pre-training and fine-tuning stages, and has officially supported 200k inference technology.

\textbf{Yi-6b-200k}~\citep{ai2024yi}: Yi series models are the next generation of open-source large language models trained from scratch by 01.AI and the 6B version is open-sourced and available to the public in November 2023 and supports a context window length of 200k.

\textbf{Yarn-Mistral-7b-128k}~\citep{peng2023yarn}: Developed by NousResearch and released in November 2023. It is further pretrained on long context data for 1500 steps using the YaRN extension method based on Mistral-7B-v0.1 and supports a 128k token context window.

\subsection{Implementation Details}
In our implementation, the specific versions of the baseline LLMs are displayed in Table~\ref{tab:data-statistics}.
For open-source LLMs, we used the official deployment method on a single NVIDIA A100 80GB GPU. Yarn-Mistral-7b-128k and Yi-6B-200K, as base models (non-chat), accomplish tasks through text completion with some limitations in instruction and format following. The inference temperature for all LLMs is set to 0 to eliminate randomness.

We considered seven context lengths for all datasets in MedOdyssey: 4k, 8k, 16k, 32k, 64k, 128k, and 200k. In the naive needle-in-a-haystack experiment, we evaluated five depths where the needle is located: 0\%, 25\%, 50\%, 75\%, and 100\%.

In MedOdyssey, almost all ground truths are context-based and close-ended. We adopted a mainstream prompting approach, providing clear task definitions and inputs/outputs, and asking LLMs to output their answers in JSON format for evaluation, the specific prompts are shown in Appendix Figure~\ref{fig:task_niah_prompt_example} to~\ref{fig:task_case_prompt_example}. Meanwhile, We specify in Table~\ref{tab:data-statistics} that the evaluation metrics for each task vary according to different ground truths. All metrics in the main experiment are computed based on exact string matching (ESM).

\begin{table}[ht]
    \centering \small
    \caption{The main experiment result of the En.Counting and Zh.Counting tasks.}
    \resizebox{0.8\linewidth}{!}{
      \begin{tabular}{l|cccccc|ccccccc}
        \toprule
        \multirow{2}{*}{Models} & \multicolumn{5}{c}{\textbf{En.Counting}} & \multirow{2}{*}{All} & \multicolumn{5}{c}{\textbf{Zh.Counting}} & \multirow{2}{*}{All} \\
        \cmidrule(lr){3-6}\cmidrule(lr){9-12}
        &                       & Rep. & Inc. & Shuf. & Cor. &  &              & Rep. & Inc. & Shuf. & Cor. & \\ \midrule
        GPT-4                   & & 0    & 5    & 1     & 1     & 7$/$28      &    & 0    & 6    & 2     & 0     & 8$/$28  \\
        \rowcolor{blue!8}GPT-4o & & 1    & 5    & 3     & 0     & \textbf{9$/$28}      &    & 1    & 6    & 3     & 0     & \textbf{10$/$28} \\
        Claude 3 Haiku          & & 0    & 7    & 1     & 0     & 8$/$28      &    & 0    & 6    & 1     & 0     & 7$/$28  \\
        Claude 3 Sonnet         & & 1    & 6    & 1     & 0     & 8$/$28      &    & 0    & 3    & 1     & 0     & 4$/$28  \\
        Moonshot-v1             & & 0    & 5    & 1     & 0     & 6$/$28      &    & 0    & 6    & 1     & 0     & 7$/$28  \\ \midrule
        ChatGLM3-6b-128k        & & 0    & 1    & 0     & 0     & 1$/$28      &    & 0    & 0    & 0     & 0     & 0$/$28  \\
        InternLM2-chat-7b       & & 0    & 1    & 1     & 0     & 2$/$28      &    & 0    & 2    & 0     & 0     & 2$/$28  \\
        InternLM2-chat-20b      & & 0    & 2    & 0     & 0     & 2$/$28      &    & 0    & 3    & 0     & 0     & 3$/$28  \\ \midrule
        \rowcolor{green!8}Yi-6b-200k              & & 0    & 0    & 0     & 0     & 0$/$28      &    & 0    & 0    & 0     & 0     & 0$/$28  \\ 
        \rowcolor{green!8}Yarn-Mistral-7b-128k    & & 0    & 0    & 0     & 0     & 0$/$28      &    & 0    & 0    & 0     & 0     & 0$/$28 \\ 
        \bottomrule
      \end{tabular}}
\label{tab:counting_ex}
\end{table}
\subsection{Results and Analysis}
\textbf{NIAH Results and Analysis.}
Figure~\ref{fig:niah_heatmap} shows the results of the naive medical-context needle-in-a-haystack experiment, using heatmaps to illustrate the performance of LLMs at different lengths and depths. We selected three representative models: GPT-4o, Claude 3 Haiku, and InternLM2-chat-20b, and the complete experimental results are shown in Appendix Table~\ref{tab:needles_ex} and Figure~\ref{fig:niah_result_all}.

Advanced LLMs, like the GPT-4 series, perform well on the naive needle-in-a-haystack task, even with new facts in the inserted needle. In contrast, other competitive LLMs see degraded performance as context length increases. Most open-source models score zero due to their inability to format outputs correctly for lengthy texts, especially the two foundational models. To address this, we relaxed the evaluation standard by removing formatting and using the subset string matching (SSM) algorithm, with results shown in Appendix Table~\ref{tab:needles_ex_ssm} and Figure~\ref{fig:niah_result_all_ssm}. Additionally, our error analysis showed that within the medical context, LLMs are more likely to make mistakes when addressing general ``needles'' compared to medical-specific ``needles'', with the error ratio being approximately 6:5.

\textbf{Counting Results and Analysis.}
We present the performance of LLMs on four types of different Counting tasks in detail in Table~\ref{tab:counting_ex} and an intuitive bar chart in Figure~\ref{fig:counting_result}. This task is quite difficult with its fictional, counter-intuitive setting, even when using state-of-the-art LLMs. There is an interesting phenomenon where advanced LLMs can perform increasing counting tasks, likely due to their ability to capture this incremental pattern from the training corpus. However, this ability fades with disorganized counting. Most LLMs struggle with repeated counting and counting with corrections, highlighting their diminished reasoning ability, similar to a student confused by similar answer choices. Additionally, it reveals their vulnerability to self-doubt, akin to a student who becomes skeptical when all answer options are identical.

\begin{figure}[ht]
    \centering
    \includegraphics[width=\linewidth]{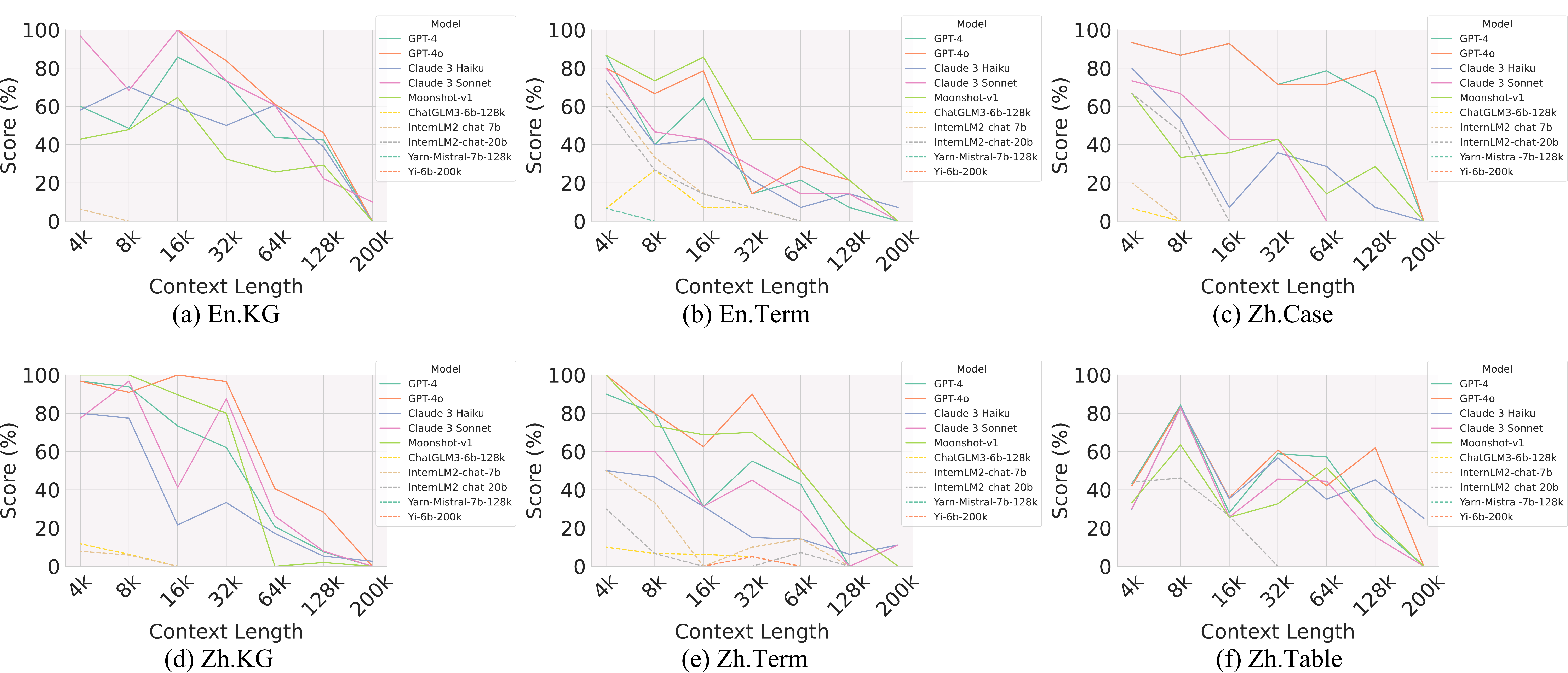}
    \caption{Trends in the performance variations of LLMs on medical-related tasks across different context lengths.}
    \label{fig:task_result_per_sample_size}
\end{figure}

\begin{table}[ht]
    \centering 
    \small
    \caption{The main experiment results of medical-related tasks based on exact string matching.}
    \resizebox{1.0\linewidth}{!}{
      \begin{tabular}{@{}lccccccccccccc@{}}
        \toprule
        \multicolumn{1}{l}{\multirow{2}{*}{Models}}            & \multicolumn{3}{c}{\textbf{En.KG}} & \multicolumn{3}{c}{\textbf{Zh.KG}}     & \textbf{En.Term} & \textbf{Zh.Term} & \textbf{Zh.Case} & \multicolumn{3}{c}{\textbf{Zh.Table}} \\
        \cmidrule(lr){2-4}\cmidrule(lr){5-7}\cmidrule(lr){8-8}\cmidrule(lr){9-9}\cmidrule(lr){10-10}\cmidrule(lr){11-13}
                                                               & P.    & R.    & F1.   & P.    & R.    & F1.   & Acc.  & Acc.  & Acc.  & P.    & R.    & F1.  & \\ \midrule
        GPT-4                                                  & 59.34 & 47.37 & 52.68 & 42.28 & 31.03 & 35.80 & 34.00 & 43.00 & 70.00 & 46.27 & 44.29 & 45.26 \\
        \rowcolor{blue!8}GPT-4o                                & \textbf{76.70} & \textbf{69.30} & \textbf{72.81} & \textbf{76.58} & \textbf{41.87} & \textbf{54.14} & 42.00 & \textbf{60.00} & \textbf{71.00} & \textbf{48.00} & \textbf{51.43} & \textbf{49.66} \\
        Claude 3 Haiku                                         & 53.54 & 46.49 & 49.77 & 21.19 & 24.63 & 22.78 & 30.00 & 24.00 & 31.00 & 45.86 & 43.57 & 44.69 \\
        Claude 3 Sonnet                                        & 72.04 & 58.77 & 64.73 & 48.39 & 29.56 & 36.70 & 33.00 & 34.00 & 33.00 & 39.55 & 37.86 & 38.69 \\
        Moonshot-v1                                            & 33.33 & 42.11 & 37.21 & 62.07 & 26.60 & 37.24 & \textbf{51.00} & 56.00 & 32.00 & 36.15 & 34.31 & 35.21 \\ \midrule
        ChatGLM3-6b-128k                                       & 0.00  & 0.00  &  0.00 & 7.89  & 1.48  & 2.49  & 7.00  & 4.00  & 1.00  & 0.00  & 0.00  & 0.00 \\
        InternLM2-chat-7b  & 2.90  & 1.75  & 2.19  & 5.45  & 1.48  & 2.33  & 18.00 & 14.00 & 3.00  & 0.00  & 0.00  & 0.00 \\
        InternLM2-chat-20b & 0.00  & 0.00  & 0.00  & 0.00  & 0.00  & 0.00  & 16.00 & 5.00  & 17.00 & 31.63 & 22.14 & 26.05 \\ \midrule
        \rowcolor{green!8}Yi-6b-200k                           & 0.00  & 0.00  & 0.00  & 0.00  & 0.00  & 0.00  & 0.00  & 1.00  & 0.00  & 0.00  & 0.00  & 0.00 \\ 
        \rowcolor{green!8}Yarn-Mistral-7b-128k                 & 0.00  & 0.00  & 0.00  & 0.00  & 0.00  & 0.00  & 1.00  & 0.00  & 0.00  & 0.00  & 0.00  & 0.00 \\ 
        \bottomrule
      \end{tabular}}
\label{tab:task_ex}
\end{table}
\textbf{Medical-related Tasks Results and Analysis.}
The overall performance of medical-related tasks is displayed in Table~\ref{tab:task_ex}, and in the meantime we provide a loose version of the results using SSM in Table~\ref{tab:task_ex_ssm}. The current state-of-the-art GPT-4o model performs well in terms of answer quality and format adherence, but is still not entirely reliable. Notably, the model's performance exhibits an overall decline as the context length increases, as shown in Figure~\ref{fig:task_result_per_sample_size}. The open-source LLMs are almost impossible to accomplish the task, especially two base models, which lose the ability to output in format (marked with a green background). In particular, Moonshot-v1 has a good performance if only the content of the answer is considered for evaluation.

\begin{figure}[ht]
    \centering
    \includegraphics[width=\linewidth]{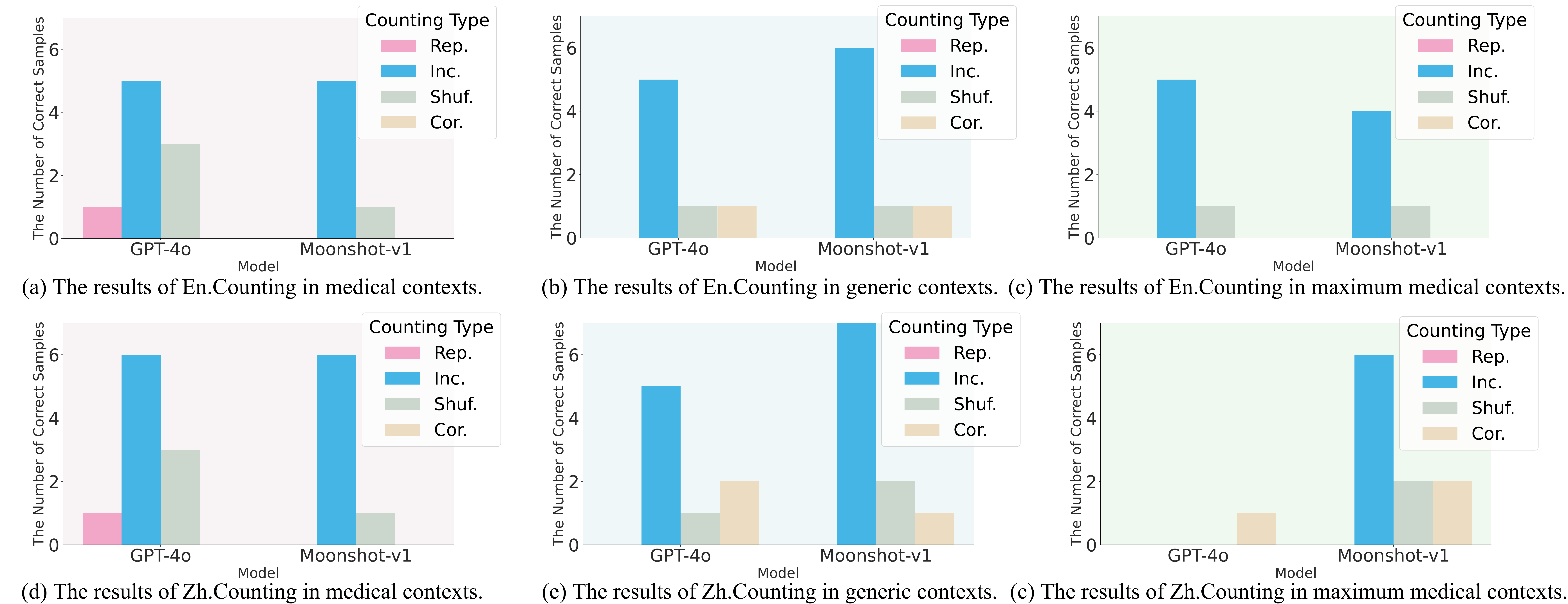}
    \caption{Comparison of GPT-4o and Moonshot-v1 on Counting tasks in different context settings.}
    \label{fig:counting_compare_result}
\end{figure}

\textbf{Analysis of Different Context Setting.}
We used the Counting task to experiment with different context settings: medical long context (MIC), generic long context (MIC), and maximum medical context length. The ablation results are shown in Figure~\ref{fig:counting_compare_result}. The experimental results support our proposed ``MIC'' principle. It is easy to observe that the performance is affected by different contexts whether the length is different or the domain is different, so we prefer to sacrifice an evaluation of extreme context length in exchange for sharing the same contextual texts between different LLMs. Due to different training corpus and training strategies, the degree of impact varies.

\section{Conclusion and Limitation}
\label{sec:con_limit}
We take a step forward by building the first medical-domain long-context evaluation benchmark, \textbf{MedOdyssey}, to facilitate the study of LLMs in long-context scenarios. Our benchmarks include medical-context needle-in-a-haystack tasks and several medical-related long-context tasks, spanning ten evaluation datasets. Additionally, we propose three effective principles to enhance the fairness and reliability of evaluations. We assess ten state-of-the-art LLMs, providing performance results and analyses in various formats. Additionally, we provide examples of the impact of different contexts.

Medical long-context evaluation is challenging, and our work faces some dilemmas. We sacrificed evaluating limit lengths to ensure different models share the same contextual cues, resulting in a restricted length being assessed. Effective open-ended QA is lacking due to difficulty in finding appropriate evaluation methods. Additionally, we took efforts to eliminate the effects of randomness (by fixing temperature and format constraints) and prevent data leakage, but these issues are unavoidable. We will continuously explore ways to improve our benchmark, as mentioned in Section~\ref{sec: construction}.
\section{Ethical Considerations}
\label{sec:ethics}
This paper proposes a new medical-domain long-context evaluation benchmark \textbf{MedOdyssey} for LLMs. All of the datasets in MedOdyssey are adhere to ethical guidelines and respect copyright laws. The entire data collection process is free of issues of copyright and issues of privacy, and there are three types of data sources, including license applications, the open source community, and public file cleaning and organizing. Meanwhile, the manual participation part in the dataset construction process was all done by the authors of this paper without any ethical issues.

\bibliographystyle{plainnat}
\bibliography{MedOdyssey_arxiv}
\section*{Checklist}

The checklist follows the references.  Please
read the checklist guidelines carefully for information on how to answer these
questions.  For each question, change the default \answerTODO{} to \answerYes{},
\answerNo{}, or \answerNA{}.  You are strongly encouraged to include a {\bf
justification to your answer}, either by referencing the appropriate section of
your paper or providing a brief inline description.  For example:
\begin{itemize}
  \item Did you include the license to the code and datasets? \answerYes
  \item Did you include the license to the code and datasets? \answerNo{The code and the data are proprietary.}
  \item Did you include the license to the code and datasets? \answerNA{}
\end{itemize}
Please do not modify the questions and only use the provided macros for your
answers.  Note that the Checklist section does not count towards the page
limit.  In your paper, please delete this instructions block and only keep the
Checklist section heading above along with the questions/answers below.

\appendix

\renewcommand{\thetable}{A\arabic{table}}
\renewcommand{\thefigure}{A\arabic{figure}}
\setcounter{figure}{0}
\setcounter{table}{0}
\clearpage

\section{Supplementary material for the experiment}
\begin{table}[ht]
    \centering \small
    \caption{The token-to-character conversion ratios of advanced long-context LLMs.}
	\resizebox{1.0\linewidth}{!}{
	\begin{tabular}{lcccccccc}
		\toprule
		\textbf{Models} & \textbf{En.NIAH} & \textbf{Zh.NIAH} & \textbf{En.KG} & \textbf{Zh.KG} & \textbf{En.Term} & \textbf{Zh.Term} & \textbf{Zh.Case} & \textbf{Zh.Table} \\
		\midrule
		  GPT-4 & 0.281 & \cellcolor{blue!8}\textbf{1.402} & 0.267 & \cellcolor{blue!8}\textbf{1.473} & 0.267 & \cellcolor{blue!8}\textbf{1.446$-$1.676} & \cellcolor{blue!8}\textbf{1.316} & \cellcolor{blue!8}\textbf{1.178} \\
		GPT-4o & 0.275 & 1.005 & 0.252 & 1.029 & 0.253 & 0.991$-$1.124 & 0.904 & 0.802  \\
            Claude 3 Haiku$/$Sonnet & 0.289 & 1.342 & 0.275 & 1.330 & 0.264 & 1.291$-$1.483 & 1.191 & 1.072 \\
            Moonshot-v1 & 0.286 & 0.924 & 0.266 & 0.737 & 0.265 & 0.732$-$0.780 & 0.712 & 0.600  \\
            \midrule
            ChatGLM3-6b-128k & 0.342 & 0.924 & 0.313 & 0.750 & 0.302 & 0.760$-$0.827 & 0.746 & 0.630 \\
            InternLM2-chat-7b$/$20b & 0.299 & 0.899 & 0.292 & 0.739 & 0.289 & 0.750$-$0.797 & 0.725 & 0.608 \\
            Yi-6b-200k & 0.342 & 0.992 & 0.301 & 0.812 & 0.293 & 0.791$-$0.883 & 0.773 & 0.659 \\
            Yarn-Mistral-7b-128k & \cellcolor{blue!8}\textbf{0.355} & 1.394 & \cellcolor{blue!8}\textbf{0.331} & 1.430 & \cellcolor{blue!8}\textbf{0.324} & 1.362$-$1.607 & 1.286 & 1.139 \\
		\bottomrule
    \end{tabular}}
\label{tab:token_scale_ratio}
\end{table}

\begin{table}[ht]
    \centering 
    \small
    \caption{The main experiment results of NIAH.}
    \resizebox{1.0\linewidth}{!}{
      \begin{tabular}{@{}l|c|cccccccc|cccccccc@{}}
        \toprule
        \multirow{2}{*}{Models} &{\multirow{2}{*}{Ablation}} & \multicolumn{7}{c}{\textbf{En.NIAH}} & \multirow{2}{*}{ALL} & \multicolumn{7}{c}{\textbf{Zh.NIAH}} & \multirow{2}{*}{ALL} \\
        \cmidrule(lr){3-9}\cmidrule(lr){11-17}
                                              &         & 4k       & 8k       & 16k      & 32k      & 64k      & 128k     & 200k &     & 4k   & 8k   & 16k  & 32k  & 64k  & 128k & 200k \\ 
        \midrule
        \midrule
        \multirow{6}{*}{GPT-4}                & 0$\%$   & 19 & 17 & 18 & 18 & 18 & 17 & $-$ & 107$/$120 & 19 & 19 & 18 & 18 & 18 & 17 & $-$ & 109$/$120 \\
                                              & 25$\%$  & 18 & 19 & 18 & 18 & 15 & 14 & $-$ & 102$/$120 & 19 & 18 & 19 & 18 & 19 & 19 & $-$ & 112$/$120 \\
                                              & 50$\%$  & 16 & 18 & 17 & 17 & 16 & 16 & $-$ & 100$/$120 & 18 & 18 & 19 & 19 & 18 & 18 & $-$ & 110$/$120 \\ 
                                              & 75$\%$  & 16 & 18 & 18 & 19 & 18 & 15 & $-$ & 104$/$120 & 20 & 18 & 19 & 19 & 18 & 18 & $-$ & 112$/$120 \\ 
                                              & 100$\%$ & 18 & 17 & 16 & 18 & 16 & 16 & $-$ & 101$/$120 & 19 & 19 & 20 & 20 & 20 & 18 & $-$ & 116$/$120 \\ 
                                              & ALL     & 87$/$100 & 89$/$100 & 87$/$100 & 90$/$100 & 83$/$100 & 78$/$100 & $-$ & 514$/$600 & 95$/$100 & 92$/$100 & 95$/$100 & 94$/$100 & 93$/$100 & 90$/$100 & $-$ & 559$/$600 \\ \midrule
        \multirow{6}{*}{GPT-4o}               & 0$\%$   & 16 & 15 & 16 & 17 & 16 & 16 & $-$ & 96$/$120 & 19 & 19 & 16 & 19 & 19 & 19 & $-$ & 111$/$120 \\
                                              & 25$\%$  & 16 & 15 & 17 & 18 & 17 & 15 & $-$ & 98$/$120 & 19 & 18 & 17 & 18 & 19 & 17 & $-$ & 108$/$120 \\
                                              & 50$\%$  & 16 & 16 & 17 & 17 & 17 & 16 & $-$ & 99$/$120 & 19 & 19 & 18 & 19 & 19 & 17 & $-$ & 111$/$120 \\
                                              & 75$\%$  & 16 & 17 & 17 & 16 & 16 & 17 & $-$ & 99$/$120 & 16 & 19 & 18 & 19 & 17 & 19 & $-$ & 108$/$120 \\
                                              & 100$\%$ & 17 & 18 & 18 & 19 & 18 & 16 & $-$ & 106$/$120 & 18 & 19 & 18 & 19 & 19 & 19 & $-$ & 112$/$120 \\
                                              & ALL     & 81$/$100 & 81$/$100 & 85$/$100 & 87$/$100 & 84$/$100 & 80$/$100 & $-$ & 498$/$600 & 91$/$100 & 94$/$100 & 87$/$100 & 94$/$100 & 93$/$100 & 91$/$100 & $-$ & 550$/$600 \\ \midrule
        \multirow{6}{*}{Claude 3 Haiku}       & 0$\%$   & 7 & 6 & 4 & 6 & 6 & 2 & 0 & 31$/$140 & 9 & 11 & 6 & 8 & 5 & 3 & 7 & 49$/$140 \\
                                              & 25$\%$  & 6 & 5 & 0 & 0 & 1 & 0 & 0 & 12$/$140 & 9 & 9 & 9 & 8 & 8 & 0 & 4 & 47$/$140 \\
                                              & 50$\%$  & 7 & 3 & 1 & 1 & 2 & 0 & 0 & 14$/$140 & 10 & 7 & 8 & 6 & 6 & 0 & 6 & 43$/$140 \\
                                              & 75$\%$  & 8 & 2 & 1 & 0 & 3 & 0 & 0 & 14$/$140 & 7 & 9 & 8 & 5 & 7 & 0 & 4 & 40$/$140 \\
                                              & 100$\%$ & 7 & 2 & 0 & 0 & 1 & 0 & 0 & 10$/$140 & 10 & 9 & 2 & 2 & 4 & 0 & 8 & 35$/$140 \\
                                              & ALL     & 35$/$100 & 18$/$100 & 6$/$100 & 7$/$100 & 13$/$100 & 2$/$100 & 0$/$100 & 81$/$700 & 45$/$100 & 45$/$100 & 33$/$100 & 29$/$100 & 30$/$100 & 3$/$100 & 29$/$100 & 214$/$700 \\ \midrule
        \multirow{6}{*}{Claude 3 Sonnet}      & 0$\%$   & 7 & 6 & 9 & 5 & 1 & 0 & 0 & 28$/$140 & 9 & 5 & 5 & 3 & 9 & 0 & 0 & 31$/$140 \\
                                              & 25$\%$  & 3 & 4 & 1 & 0 & 0 & 0 & 0 & 8$/$140 & 8 & 5 & 4 & 4 & 0 & 0 & 0 & 21$/$140 \\
                                              & 50$\%$  & 5 & 1 & 0 & 1 & 0 & 0 & 0 & 7$/$140 & 5 & 5 & 4 & 5 & 0 & 0 & 0 & 19$/$140 \\
                                              & 75$\%$  & 7 & 2 & 1 & 0 & 0 & 0 & 0 & 10$/$140 & 7 & 4 & 4 & 3 & 0 & 0 & 0 & 18$/$140 \\
                                              & 100$\%$ & 2 & 3 & 0 & 0 & 0 & 0 & 0 & 5$/$140 & 5 & 3 & 5 & 2 & 0 & 0 & 0 & 15$/$140 \\
                                              & ALL     & 24$/$100 & 16$/$100 & 11$/$100 & 6$/$100 & 1$/$100 & 0$/$100 & 0$/$100 & 58$/$700 & 34$/$100 & 22$/$100 & 22$/$100 & 17$/$100 & 9$/$100 & 0$/$100 & 0$/$100 & 104$/$700 \\ \midrule
        \multirow{6}{*}{Moonshot-v1}          & 0$\%$   & 17 & 18 & 17 & 17 & 16 & 6 & $-$ & 91$/$120 & 16 & 16 & 11 & 7 & 2 & 1 & $-$ & 53$/$120 \\
                                              & 25$\%$  & 17 & 15 & 14 & 12 & 10 & 2 & $-$ & 70$/$120 & 16 & 16 & 6 & 3 & 1 & 1 & $-$ & 43$/$120 \\
                                              & 50$\%$  & 16 & 17 & 14 & 10 & 7 & 4 & $-$ & 68$/$120 & 16 & 16 & 9 & 4 & 1 & 0 & $-$ & 46$/$120 \\
                                              & 75$\%$  & 16 & 16 & 14 & 9 & 10 & 2 & $-$ & 67$/$120 & 16 & 15 & 6 & 4 & 2 & 0 & $-$ & 43$/$120 \\
                                              & 100$\%$ & 16 & 17 & 16 & 11 & 9 & 8 & $-$ & 77$/$120 & 17 & 16 & 12 & 8 & 2 & 2 & $-$ & 57$/$120 \\
                                              & ALL     & 82$/$100 & 83$/$100 & 75$/$100 & 59$/$100 & 52$/$100 & 22$/$100 & $-$ & 373$/$600 & 81$/$100 & 79$/$100 & 44$/$100 & 26$/$100 & 8$/$100 & 4$/$100 & $-$& 242$/$600 \\ \midrule
        \midrule
        \multirow{6}{*}{ChatGLM3-6b-128k}     & 0$\%$   & 1 & 0 & 0 & 0 & 0 & 0 & $-$ & 1$/$120 & 8 & 0 & 2 & 0 & 0 & 0 & $-$ & 10$/$120 \\
                                              & 25$\%$  & 0 & 0 & 0 & 0 & 0 & 0 & $-$ & 0$/$120 & 4 & 0 & 0 & 0 & 0 & 0 & $-$ & 4$/$120 \\
                                              & 50$\%$  & 0 & 0 & 0 & 0 & 0 & 0 & $-$ & 0$/$120 & 2 & 0 & 0 & 0 & 0 & 0 & $-$ & 2$/$120 \\
                                              & 75$\%$  & 0 & 0 & 0 & 0 & 0 & 0 & $-$ & 0$/$120 & 1 & 0 & 0 & 0 & 0 & 0 & $-$ & 1$/$120 \\
                                              & 100$\%$ & 0 & 0 & 0 & 0 & 0 & 0 & $-$ & 0$/$120 & 4 & 0 & 2 & 0 & 0 & 0 & $-$ & 6$/$120 \\
                                              & ALL     & 1$/$100 & 0$/$100 & 0$/$100 & 0$/$100 & 0$/$100 & 0$/$100 & $-$ & 1$/$600 & 19$/$100 & 0$/$100 & 4$/$100 & 0$/$100 & 0$/$100 & 0$/$100 & $-$ & 23$/$600 \\ \midrule
        \multirow{6}{*}{InternLM2-chat-7b}    & 0$\%$   & 11 & 3 & 0 & 0 & 0 & 0 & 0 & 14$/$140 & 4 & 4 & 0 & 0 & 0 & 0 & 0 & 8$/$140 \\
                                              & 25$\%$  & 11 & 2 & 0 & 0 & 0 & 0 & 0 & 13$/$140 & 5 & 2 & 0 & 0 & 0 & 0 & 0 & 7$/$140 \\
                                              & 50$\%$  & 8 & 3 & 0 & 0 & 0 & 0 & 0 & 11$/$140 & 1 & 1 & 0 & 0 & 0 & 0 & 0 & 2$/$140 \\
                                              & 75$\%$  & 6 & 3 & 0 & 0 & 0 & 0 & 0 & 9$/$140 & 0 & 0 & 0 & 0 & 0 & 0 & 0 & 0$/$140 \\
                                              & 100$\%$ & 7 & 2 & 0 & 0 & 0 & 0 & 0 & 9$/$140 & 0 & 0 & 0 & 0 & 0 & 0 & 0 & 0$/$140 \\
                                              & ALL     & 43$/$100 & 13$/$100 & 0$/$100 & 0$/$100 & 0$/$100 & 0$/$100 & 0$/$100 & 56$/$700 & 10$/$100 & 7$/$100 & 0$/$100 & 0$/$100 & 0$/$100 & 0$/$100 & 0$/$100 & 17$/$700 \\ \midrule
        \multirow{6}{*}{InternLM2-chat-20b}   & 0$\%$   & 12 & 11 & 0 & 0 & 0 & 0 & 0 & 23$/$140 & 16 & 14 & 12 & 2 & 0 & 0 & 0 & 44$/$140 \\
                                              & 25$\%$  & 12 & 8 & 0 & 0 & 0 & 0 & 0 & 20$/$140 & 14 & 12 & 6 & 0 & 0 & 0 & 0 & 32$/$140 \\
                                              & 50$\%$  & 9 & 8 & 0 & 0 & 0 & 0 & 0 & 17$/$140 & 13 & 11 & 6 & 1 & 0 & 0 & 0 & 31$/$140 \\
                                              & 75$\%$  & 11 & 8 & 0 & 0 & 0 & 0 & 0 & 19$/$140 & 13 & 10 & 7 & 0 & 0 & 0 & 0 & 30$/$140 \\
                                              & 100$\%$ & 14 & 10 & 0 & 0 & 0 & 0 & 0 & 24$/$140 & 14 & 13 & 8 & 2 & 0 & 0 & 0 & 37$/$140 \\
                                              & ALL     & 58$/$100 & 45$/$100 & 0$/$100 & 0$/$100 & 0$/$100 & 0$/$100 & 0$/$100 & 103$/$700 & 70$/$100 & 60$/$100 & 39$/$100 & 5$/$100 & 0$/$100 & 0$/$100 & 0$/$100 & 174$/$700 \\ \midrule
        \midrule
        \multirow{6}{*}{Yarn-Mistral-7b-128k} & 0$\%$   & 0 & 0 & 0 & 0 & 0 & 0 & $-$ & 0$/$120 & 1 & 0 & 0 & 0 & 0 & 0 & $-$ & 1$/$120 \\
                                              & 25$\%$  & 0 & 0 & 0 & 0 & 0 & 0 & $-$ & 0$/$120 & 0 & 0 & 0 & 0 & 0 & 0 & $-$ & 0$/$120 \\
                                              & 50$\%$  & 0 & 0 & 0 & 0 & 0 & 0 & $-$ & 0$/$120 & 0 & 0 & 0 & 0 & 0 & 0 & $-$ & 0$/$120 \\
                                              & 75$\%$  & 0 & 0 & 0 & 0 & 0 & 0 & $-$ & 0$/$120 & 0 & 0 & 0 & 0 & 0 & 0 & $-$ & 0$/$120 \\
                                              & 100$\%$ & 0 & 0 & 0 & 0 & 0 & 0 & $-$ & 0$/$120 & 3 & 0 & 0 & 0 & 0 & 0 & $-$ & 3$/$120 \\
                                              & ALL     & 0$/$100 & 0$/$100 & 0$/$100 & 0$/$100 & 0$/$100 & 0$/$100 & $-$ & 0$/$600 & 4$/$100 & 0$/$100 & 0$/$100 & 0$/$100 & 0$/$100 & 0$/$100 & $-$ & 4$/$600 \\ \midrule
        \multirow{6}{*}{Yi-6b-200k}           & 0$\%$   & 0 & 0 & 0 & 0 & 0 & 0 & 0 & 0$/$140 & 0 & 0 & 0 & 0 & 0 & 0 & 0 & 0$/$140 \\
                                              & 25$\%$  & 0 & 0 & 0 & 0 & 0 & 0 & 0 & 0$/$140 & 0 & 0 & 0 & 0 & 0 & 0 & 0 & 0$/$140 \\
                                              & 50$\%$  & 0 & 0 & 0 & 0 & 0 & 0 & 0 & 0$/$140 & 0 & 0 & 0 & 0 & 0 & 0 & 0 & 0$/$140 \\
                                              & 75$\%$  & 0 & 0 & 0 & 0 & 0 & 0 & 0 & 0$/$140 & 0 & 0 & 0 & 0 & 0 & 0 & 0 & 0$/$140 \\
                                              & 100$\%$ & 0 & 0 & 0 & 0 & 0 & 0 & 0 & 0$/$140 & 0 & 0 & 0 & 0 & 0 & 0 & 0 & 0$/$140 \\
                                              & ALL     & 0$/$100 & 0$/$100 & 0$/$100 & 0$/$100 & 0$/$100 & 0$/$100 & 0$/$100 & 0$/$700 & 0$/$100 & 0$/$100 & 0$/$100 & 0$/$100 & 0$/$100 & 0$/$100 & 0$/$100 & 0$/$700\\ \bottomrule
      \end{tabular}}
\label{tab:needles_ex}
\end{table}

\begin{table}[ht]
    \centering 
    \small
    \caption{The main experiment result of NIAH based on subset string matching.}
    \resizebox{1.0\linewidth}{!}{
      \begin{tabular}{@{}l|c|cccccccc|cccccccc@{}}
        \toprule
        \multirow{2}{*}{Models} &{\multirow{2}{*}{Ablation}} & \multicolumn{7}{c}{\textbf{En.NIAH}} & \multirow{2}{*}{ALL} & \multicolumn{7}{c}{\textbf{Zh.NIAH}} & \multirow{2}{*}{ALL} \\
        \cmidrule(lr){3-9}\cmidrule(lr){11-17}
                                              &         & 4k       & 8k       & 16k      & 32k      & 64k      & 128k     & 200k &     & 4k   & 8k   & 16k  & 32k  & 64k  & 128k & 200k \\ 
        \midrule
        \midrule
        \multirow{6}{*}{GPT-4}                & 0$\%$   & 20 & 19 & 20 & 20 & 20 & 20 & $-$ & 119$/$120 & 19 & 19 & 18 & 18 & 18 & 18 & $-$ & 110$/$120 \\
                                              & 25$\%$  & 20 & 20 & 19 & 20 & 18 & 19 & $-$ & 116$/$120 & 19 & 18 & 19 & 18 & 19 & 20 & $-$ & 113$/$120 \\
                                              & 50$\%$  & 18 & 20 & 20 & 20 & 19 & 19 & $-$ & 116$/$120 & 19 & 18 & 19 & 19 & 19 & 18 & $-$ & 112$/$120 \\
                                              & 75$\%$  & 19 & 20 & 20 & 20 & 20 & 20 & $-$ & 119$/$120 & 20 & 19 & 19 & 19 & 19 & 19 & $-$ & 115$/$120 \\
                                              & 100$\%$ & 19 & 19 & 19 & 20 & 20 & 20 & $-$ & 117$/$120 & 19 & 19 & 20 & 20 & 20 & 19 & $-$ & 117$/$120 \\
                                              & ALL     & 96$/$100 & 98$/$100 & 98$/$100 & 100$/$100 & 97$/$100 & 98$/$100 & $-$ & 587$/$600 & 96$/$100 & 93$/$100 & 95$/$100 & 94$/$100 & 95$/$100 & 94$/$100 & $-$ & 567$/$600 \\ \midrule
        \multirow{6}{*}{GPT-4o}               & 0$\%$   & 19 & 19 & 19 & 19 & 20 & 20 & $-$ & 116$/$120 & 20 & 20 & 18 & 20 & 20 & 20 & $-$ & 118$/$120 \\
                                              & 25$\%$  & 18 & 18 & 20 & 20 & 20 & 18 & $-$ & 114$/$120 & 20 & 20 & 19 & 20 & 20 & 20 & $-$ & 119$/$120 \\
                                              & 50$\%$  & 19 & 19 & 19 & 20 & 19 & 20 & $-$ & 116$/$120 & 20 & 20 & 19 & 20 & 20 & 20 & $-$ & 119$/$120 \\
                                              & 75$\%$  & 18 & 20 & 19 & 19 & 20 & 20 & $-$ & 116$/$120 & 19 & 20 & 20 & 20 & 20 & 20 & $-$ & 119$/$120 \\
                                              & 100$\%$ & 20 & 19 & 20 & 20 & 20 & 17 & $-$ & 116$/$120 & 20 & 20 & 20 & 20 & 20 & 20 & $-$ & 120$/$120 \\
                                              & ALL     & 94$/$100 & 95$/$100 & 97$/$100 & 98$/$100 & 99$/$100 & 95$/$100 & $-$ & 578$/$600 & 99$/$100 & 100$/$100 & 96$/$100 & 100$/$100 & 100$/$100 & 100$/$100 & $-$ & 595$/$600\\ \midrule
        \multirow{6}{*}{Claude 3 Haiku}       & 0$\%$   & 19 & 20 & 20 & 20 & 20 & 16 & 18 & 133$/$140 & 19 & 20 & 19 & 19 & 19 & 18 & 18 & 132$/$140 \\
                                              & 25$\%$  & 17 & 16 & 18 & 16 & 17 & 16 & 17 & 117$/$140 & 18 & 18 & 18 & 18 & 20 & 16 & 19 & 127$/$140 \\
                                              & 50$\%$  & 16 & 19 & 20 & 17 & 17 & 19 & 19 & 127$/$140 & 19 & 20 & 18 & 19 & 19 & 19 & 19 & 133$/$140 \\
                                              & 75$\%$  & 16 & 18 & 19 & 17 & 18 & 19 & 18 & 125$/$140 & 18 & 19 & 19 & 19 & 18 & 18 & 20 & 131$/$140 \\
                                              & 100$\%$ & 18 & 20 & 19 & 19 & 19 & 19 & 19 & 133$/$140 & 18 & 19 & 18 & 19 & 18 & 19 & 19 & 130$/$140 \\
                                              & ALL     & 86$/$100 & 93$/$100 & 96$/$100 & 89$/$100 & 91$/$100 & 89$/$100 & 91$/$100 & 635$/$700 & 92$/$100 & 96$/$100 & 92$/$100 & 94$/$100 & 94$/$100 & 90$/$100 & 95$/$100 & 653$/$700 \\ \midrule
        \multirow{6}{*}{Claude 3 Sonnet}      & 0$\%$   & 18 & 19 & 19 & 19 & 16 & 13 & 13 & 117$/$140 & 19 & 19 & 20 & 19 & 19 & 18 & 17 & 131$/$140 \\
                                              & 25$\%$  & 16 & 18 & 18 & 17 & 16 & 13 & 13 & 111$/$140 & 16 & 18 & 19 & 18 & 18 & 17 & 18 & 124$/$140 \\
                                              & 50$\%$  & 15 & 18 & 17 & 18 & 17 & 15 & 14 & 114$/$140 & 15 & 17 & 19 & 18 & 19 & 19 & 19 & 126$/$140 \\
                                              & 75$\%$  & 17 & 19 & 17 & 17 & 18 & 15 & 16 & 119$/$140 & 20 & 17 & 19 & 19 & 16 & 19 & 18 & 128$/$140 \\
                                              & 100$\%$ & 18 & 20 & 17 & 19 & 18 & 18 & 17 & 127$/$140 & 19 & 17 & 19 & 19 & 18 & 18 & 18 & 128$/$140 \\
                                              & ALL     & 84$/$100 & 94$/$100 & 88$/$100 & 90$/$100 & 85$/$100 & 74$/$100 & 73$/$100 & 588$/$700 & 89$/$100 & 88$/$100 & 96$/$100 & 93$/$100 & 90$/$100 & 91$/$100 & 90$/$100 & 637$/$700 \\ \midrule
        \multirow{6}{*}{Moonshot-v1}          & 0$\%$   & 19 & 20 & 19 & 19 & 19 & 17 & $-$ & 113$/$120 & 20 & 20 & 20 & 20 & 20 & 20 & $-$ & 120$/$120 \\
                                              & 25$\%$  & 19 & 19 & 18 & 19 & 19 & 18 & $-$ & 112$/$120 & 20 & 20 & 20 & 20 & 18 & 19 & $-$ & 117$/$120 \\
                                              & 50$\%$  & 18 & 19 & 18 & 18 & 18 & 18 & $-$ & 109$/$120 & 20 & 20 & 20 & 20 & 19 & 19 & $-$ & 118$/$120 \\
                                              & 75$\%$  & 18 & 18 & 18 & 19 & 19 & 19 & $-$ & 111$/$120 & 20 & 19 & 19 & 20 & 19 & 20 & $-$ & 117$/$120 \\
                                              & 100$\%$ & 18 & 18 & 18 & 17 & 17 & 18 & $-$ & 106$/$120 & 19 & 19 & 19 & 19 & 18 & 18 & $-$ & 112$/$120 \\
                                              & ALL     & 92$/$100 & 94$/$100 & 91$/$100 & 92$/$100 & 92$/$100 & 90$/$100 & $-$ & 551$/$600 & 99$/$100 & 98$/$100 & 98$/$100 & 99$/$100 & 94$/$100 & 96$/$100 & $-$ & 584$/$600\\ \midrule
        \midrule
        \multirow{6}{*}{ChatGLM3-6b-128k}     & 0$\%$   & 17 & 18 & 17 & 17 & 18 & 16 & $-$ & 103$/$120 & 20 & 19 & 19 & 18 & 18 & 15 & $-$ & 109$/$120 \\
                                              & 25$\%$  & 17 & 17 & 18 & 18 & 16 & 14 & $-$ & 100$/$120 & 18 & 18 & 19 & 17 & 15 & 14 & $-$ & 101$/$120 \\
                                              & 50$\%$  & 17 & 17 & 17 & 18 & 15 & 15 & $-$ & 99$/$120 & 18 & 19 & 17 & 19 & 15 & 16 & $-$ & 104$/$120 \\
                                              & 75$\%$  & 17 & 15 & 18 & 17 & 17 & 19 & $-$ & 103$/$120 & 17 & 18 & 17 & 17 & 18 & 14 & $-$ & 101$/$120 \\
                                              & 100$\%$ & 15 & 16 & 14 & 16 & 15 & 16 & $-$ & 92$/$120 & 18 & 19 & 18 & 19 & 17 & 15 & $-$ & 106$/$120 \\
                                              & ALL     & 83$/$100 & 83$/$100 & 84$/$100 & 86$/$100 & 81$/$100 & 80$/$100 & $-$ & 497$/$600 & 91$/$100 & 93$/$100 & 90$/$100 & 90$/$100 & 83$/$100 & 74$/$100 & $-$ & 521$/$600 \\ \midrule
        \multirow{6}{*}{InternLM2-chat-7b}    & 0$\%$   & 20 & 19 & 19 & 17 & 17 & 12 & 1 & 105$/$140 & 19 & 19 & 19 & 19 & 16 & 13 & 5 & 110$/$140 \\
                                              & 25$\%$  & 20 & 19 & 19 & 17 & 16 & 11 & 7 & 109$/$140 & 19 & 19 & 17 & 19 & 17 & 13 & 5 & 109$/$140 \\
                                              & 50$\%$  & 20 & 19 & 19 & 17 & 14 & 8 & 12 & 109$/$140 & 19 & 19 & 18 & 17 & 13 & 10 & 6 & 102$/$140 \\
                                              & 75$\%$  & 20 & 20 & 17 & 17 & 14 & 15 & 13 & 116$/$140 & 19 & 19 & 19 & 17 & 15 & 13 & 11 & 113$/$140 \\
                                              & 100$\%$ & 20 & 20 & 19 & 18 & 19 & 18 & 10 & 124$/$140 & 19 & 19 & 19 & 19 & 19 & 19 & 15 & 129$/$140 \\
                                              & ALL     & 100$/$100 & 97$/$100 & 93$/$100 & 86$/$100 & 80$/$100 & 64$/$100 & 43$/$100 & 563$/$700 & 95$/$100 & 95$/$100 & 92$/$100 & 91$/$100 & 80$/$100 & 68$/$100 & 42$/$100 & 563$/$700 \\ \midrule
        \multirow{6}{*}{InternLM2-chat-20b}   & 0$\%$   & 20 & 19 & 19 & 16 & 14 & 8 & 4 & 100$/$140 & 19 & 19 & 18 & 18 & 14 & 9 & 8 & 105$/$140 \\
                                              & 25$\%$  & 20 & 19 & 19 & 19 & 19 & 12 & 9 & 117$/$140 & 19 & 17 & 17 & 16 & 9 & 7 & 9 & 94$/$140 \\
                                              & 50$\%$  & 20 & 19 & 19 & 19 & 15 & 17 & 16 & 125$/$140 & 18 & 18 & 18 & 18 & 12 & 7 & 8 & 99$/$140 \\
                                              & 75$\%$  & 19 & 20 & 19 & 19 & 17 & 17 & 13 & 124$/$140 & 18 & 18 & 17 & 18 & 17 & 12 & 4 & 104$/$140 \\
                                              & 100$\%$ & 19 & 19 & 19 & 20 & 19 & 18 & 16 & 130$/$140 & 18 & 18 & 18 & 18 & 19 & 17 & 16 & 124$/$140 \\
                                              & ALL     & 98$/$100 & 96$/$100 & 95$/$100 & 93$/$100 & 84$/$100 & 72$/$100 & 58$/$100 & 596$/$700 & 92$/$100 & 90$/$100 & 88$/$100 & 88$/$100 & 71$/$100 & 52$/$100 & 45$/$100 & 526$/$700 \\ \midrule
        \midrule
        \multirow{6}{*}{Yarn-Mistral-7b-128k} & 0$\%$   & 13 & 12 & 9 & 9 & 7 & 0 & $-$ & 50$/$120 & 15 & 10 & 8 & 6 & 6 & 0 & $-$ & 45$/$120 \\
                                              & 25$\%$  & 13 & 14 & 6 & 5 & 3 & 0 & $-$ & 41$/$120 & 9 & 9 & 6 & 4 & 2 & 1 & $-$ & 31$/$120 \\
                                              & 50$\%$  & 12 & 13 & 6 & 7 & 2 & 0 & $-$ & 40$/$120 & 8 & 10 & 5 & 5 & 2 & 2 & $-$ & 32$/$120 \\
                                              & 75$\%$  & 14 & 15 & 11 & 6 & 2 & 0 & $-$ & 48$/$120 & 14 & 9 & 6 & 8 & 2 & 1 & $-$ & 40$/$120 \\
                                              & 100$\%$ & 12 & 13 & 15 & 13 & 13 & 0 & $-$ & 66$/$120 & 16 & 14 & 15 & 12 & 12 & 10 & $-$ & 79$/$120 \\
                                              & ALL     & 64$/$100 & 67$/$100 & 47$/$100 & 40$/$100 & 27$/$100 & 0$/$100 & $-$ & 245$/$600 & 62$/$100 & 52$/$100 & 40$/$100 & 35$/$100 & 24$/$100 & 14$/$100 & $-$ & 227$/$600 \\ \midrule
        \multirow{6}{*}{Yi-6b-200k}           & 0$\%$   & 2 & 2 & 2 & 2 & 2 & 1 & 2 & 13$/$140 & 19 & 18 & 19 & 18 & 16 & 15 & 14 & 119$/$140 \\
                                              & 25$\%$  & 2 & 2 & 2 & 2 & 2 & 2 & 2 & 14$/$140 & 18 & 18 & 15 & 13 & 14 & 13 & 11 & 102$/$140 \\
                                              & 50$\%$  & 2 & 2 & 2 & 2 & 2 & 3 & 2 & 15$/$140 & 17 & 17 & 15 & 17 & 14 & 14 & 13 & 107$/$140 \\
                                              & 75$\%$  & 2 & 2 & 2 & 3 & 2 & 3 & 2 & 16$/$140 & 19 & 17 & 16 & 17 & 15 & 15 & 14 & 113$/$140 \\
                                              & 100$\%$ & 2 & 2 & 2 & 2 & 2 & 2 & 2 & 14$/$140 & 19 & 18 & 19 & 16 & 17 & 16 & 16 & 121$/$140 \\
                                              & ALL     & 10$/$100 & 10$/$100 & 10$/$100 & 11$/$100 & 10$/$100 & 11$/$100 & 10$/$100 & 72$/$700 & 92$/$100 & 88$/$100 & 84$/$100 & 81$/$100 & 76$/$100 & 73$/$100 & 68$/$100 & 562$/$700 \\ \bottomrule
      \end{tabular}}
\label{tab:needles_ex_ssm}
\end{table}

\begin{figure}[ht]
    \centering
    \includegraphics[width=\linewidth]{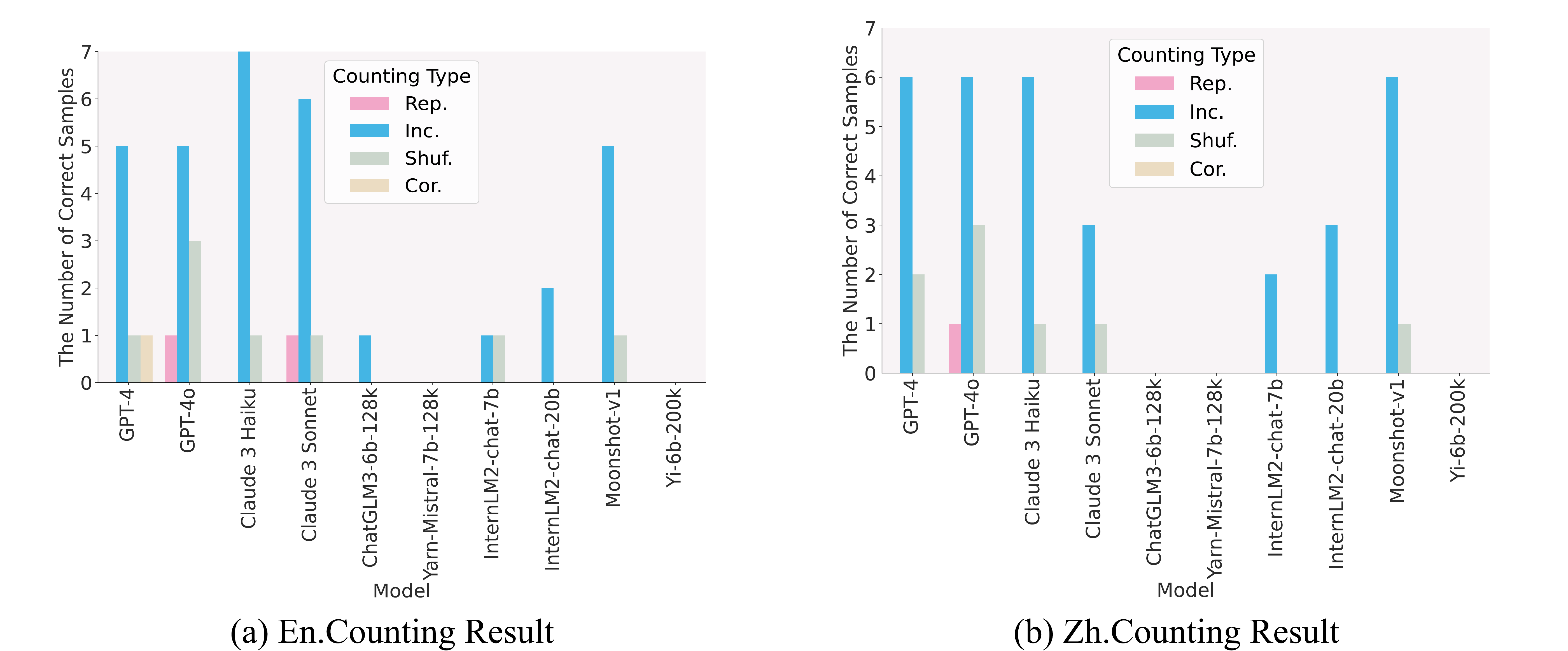}
    \caption{Histogram of Counting task results.}
    \label{fig:counting_result}
\end{figure}

\begin{figure}[ht]
    \centering
    \includegraphics[width=\linewidth]{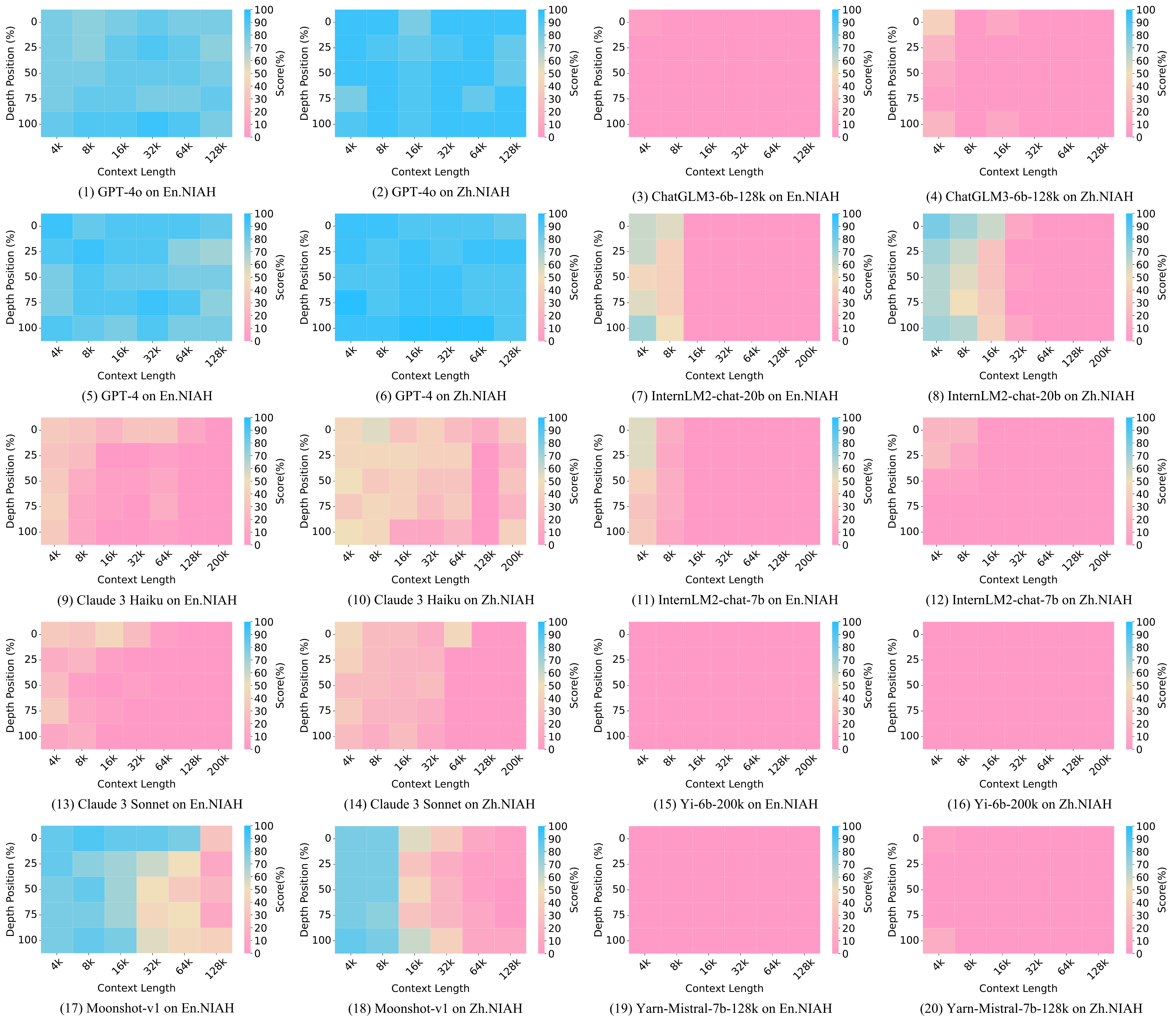}
    \caption{Heatmaps of the performance of all LLMs on NIAH task.}
    \label{fig:niah_result_all}
\end{figure}

\begin{figure}[ht]
    \centering
    \includegraphics[width=\linewidth]{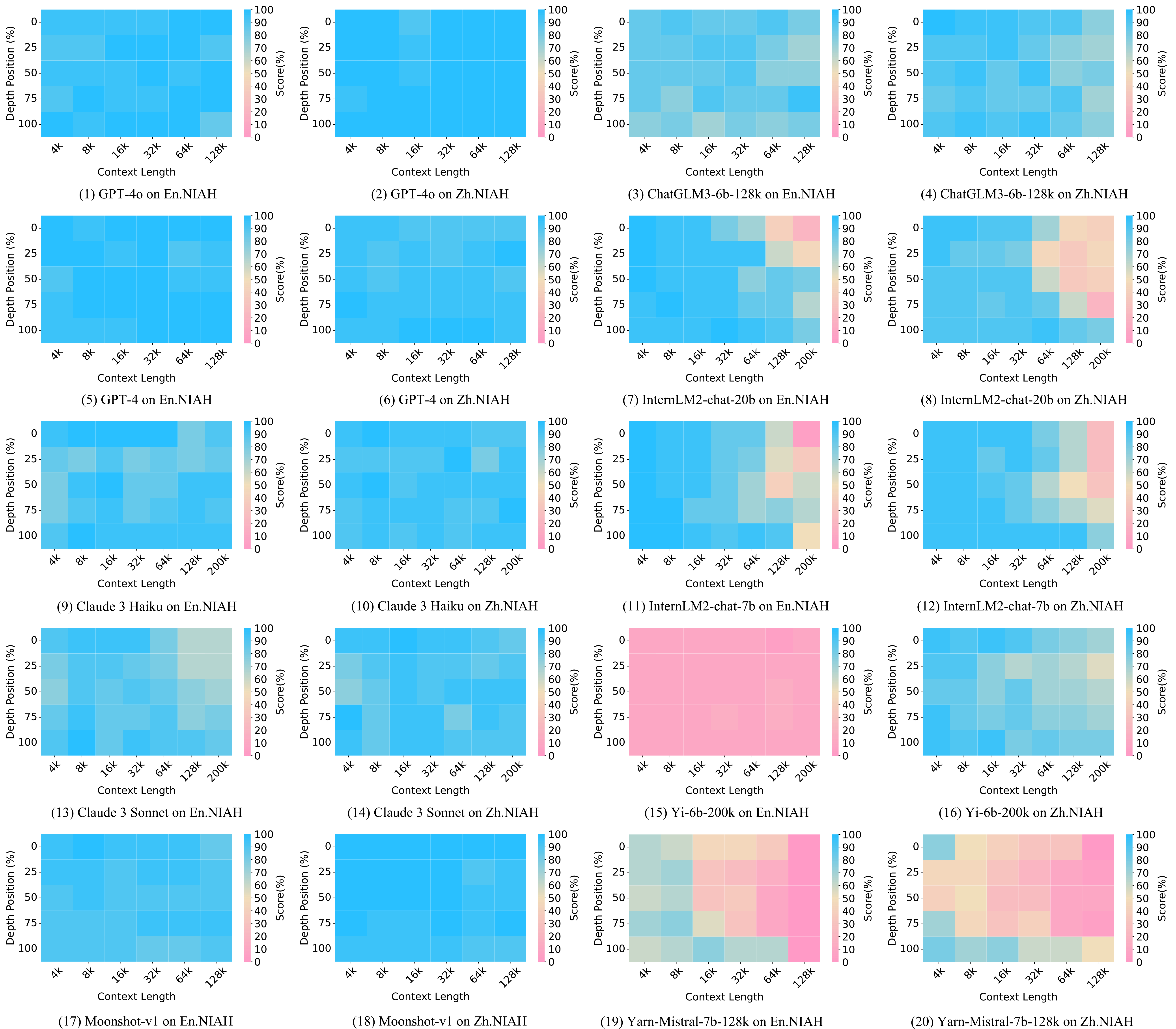}
    \caption{Heatmaps of the performance of all LLMs on NIAH task based on subset string matching.}
    \label{fig:niah_result_all_ssm}
\end{figure}

\begin{figure}[ht]
    \centering
    \includegraphics[width=\linewidth]{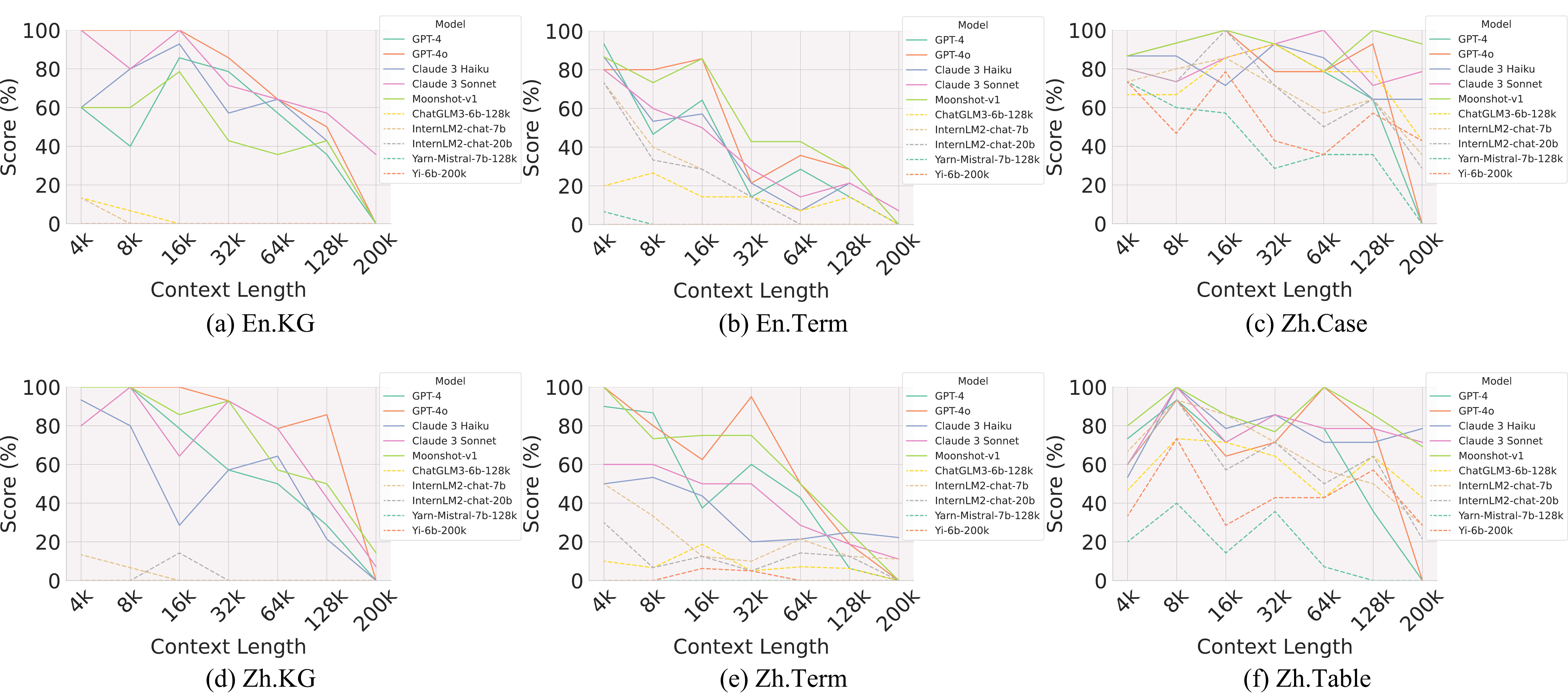}
    \caption{Trends in the performance variations of LLMs on medical-related tasks across different context lengths based on subset string matching.}
    \label{fig:task_sample_size_ssm}
\end{figure}

\begin{table}[ht]
    \centering 
    \small
    \caption{The main experiment results of medical-related tasks based on subset string matching.}
    \resizebox{1.0\linewidth}{!}{
      \begin{tabular}{@{}lccccccc@{}}
        \toprule
        Models            & \textbf{En.KG} & \textbf{Zh.KG} & \textbf{En.Term} & \textbf{Zh.Term} & \textbf{Zh.Case} & \textbf{Zh.Table} \\ \midrule
        GPT-4                                                  & 51.00 & 60.00 & 38.00 & 47.00 & 72.00 & 63.00 \\
        GPT-4o                                & 72.00 & \textbf{80.00} & 48.00 & \textbf{61.00} & 76.00 & 67.00 \\
        Claude 3 Haiku                                         & 57.00 & 50.00 & 37.00 & 33.00 & 79.00 & 77.00 \\
        Claude 3 Sonnet                                        & \textbf{73.00} & 67.00 & 38.00 & 41.00 & 83.00 & 78.00 \\
        \rowcolor{blue!8}Moonshot-v1                                            & 46.00 & 72.00 & \textbf{52.00} & 59.00 & \textbf{92.00} & \textbf{85.71} \\ \midrule
        ChatGLM3-6b-128k                                       & 3.00  & 3.00  & 14.00 & 8.00  & 73.00 & 58.00 \\
        InternLM2-chat-7b  & 2.00  & 3.00  & 23.00 & 20.00 & 67.00 & 65.00 \\
        InternLM2-chat-20b & 0.00  & 2.00  & 22.00 & 11.00 & 67.00 & 60.00 \\ \midrule
        \rowcolor{green!8}Yi-6b-200k                                             & 0.00  & 0.00  & 0.00  & 2.00  & 54.00 & 44.00 \\
        \rowcolor{green!8}Yarn-Mistral-7b-128k                                   & 0.00  & 0.00  & 1.00  & 0.00  & 42.00 & 17.00 \\ \bottomrule
      \end{tabular}}

\label{tab:task_ex_ssm}
\end{table}

\clearpage
\renewcommand{\thetable}{B\arabic{table}}
\renewcommand{\thefigure}{B\arabic{figure}}
\setcounter{figure}{0}
\setcounter{table}{0}
\section{Supplementary material for datasets}
\begin{figure}[ht]
    \centering
    \includegraphics[width=\linewidth]{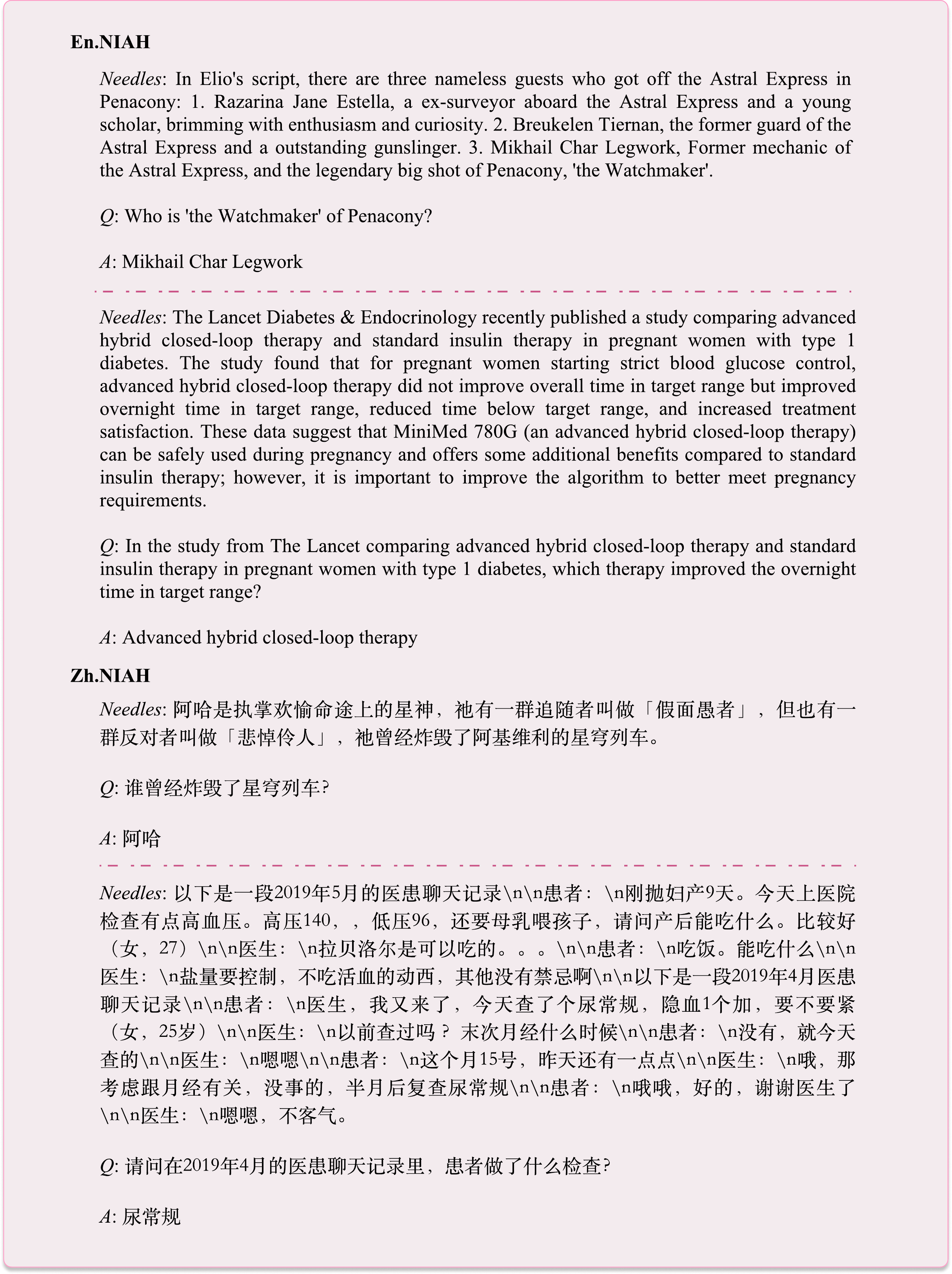}
    \caption{Examples of NIAH task.}
    \label{fig:NIAH_example}
\end{figure}

\begin{figure}[ht]
    \centering
    \includegraphics[width=\linewidth]{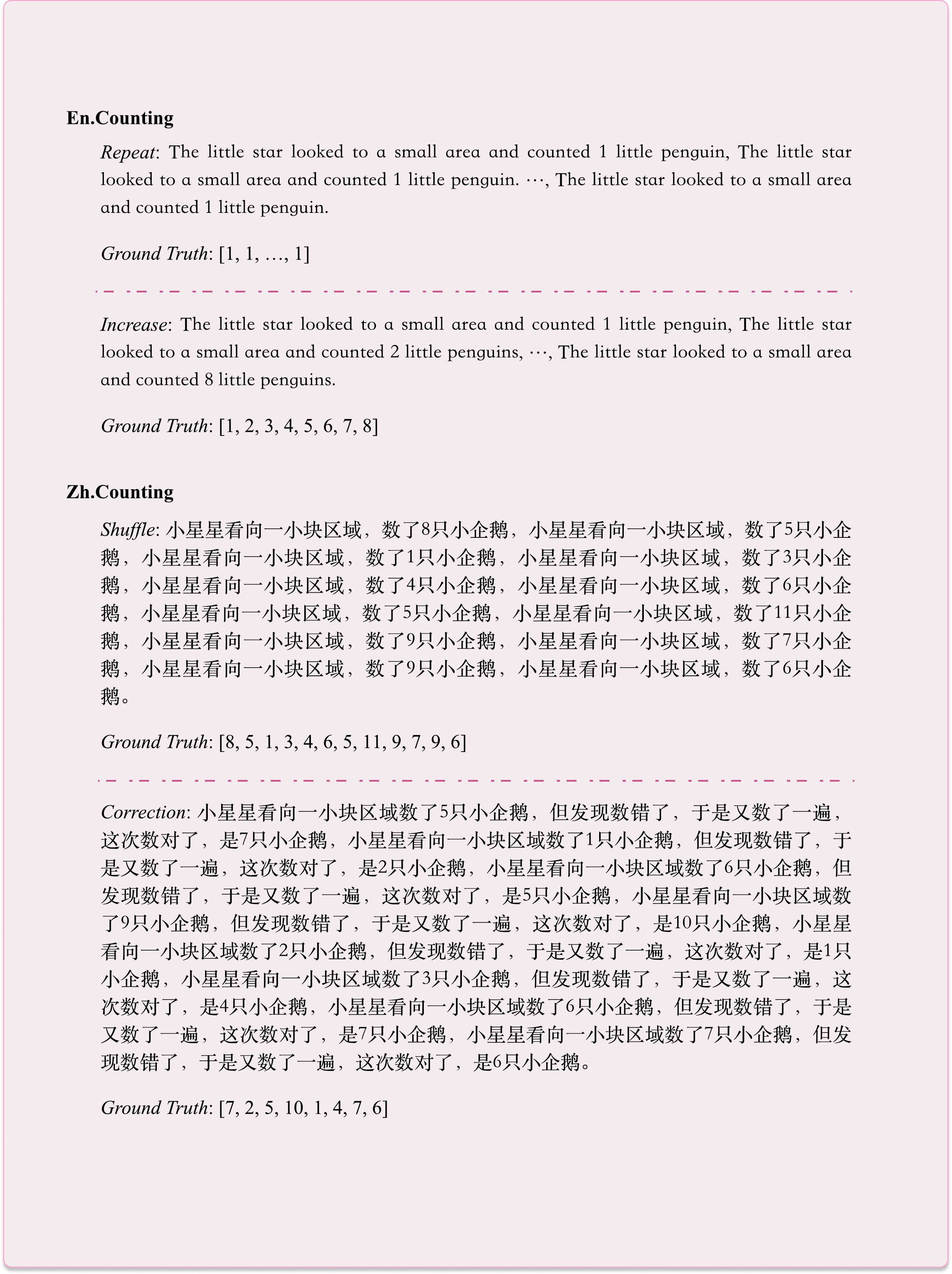}
    \caption{Examples of Counting task.}
    \label{fig:Counting_example}
\end{figure}

\begin{figure}[ht]
    \centering
    \includegraphics[width=\linewidth]{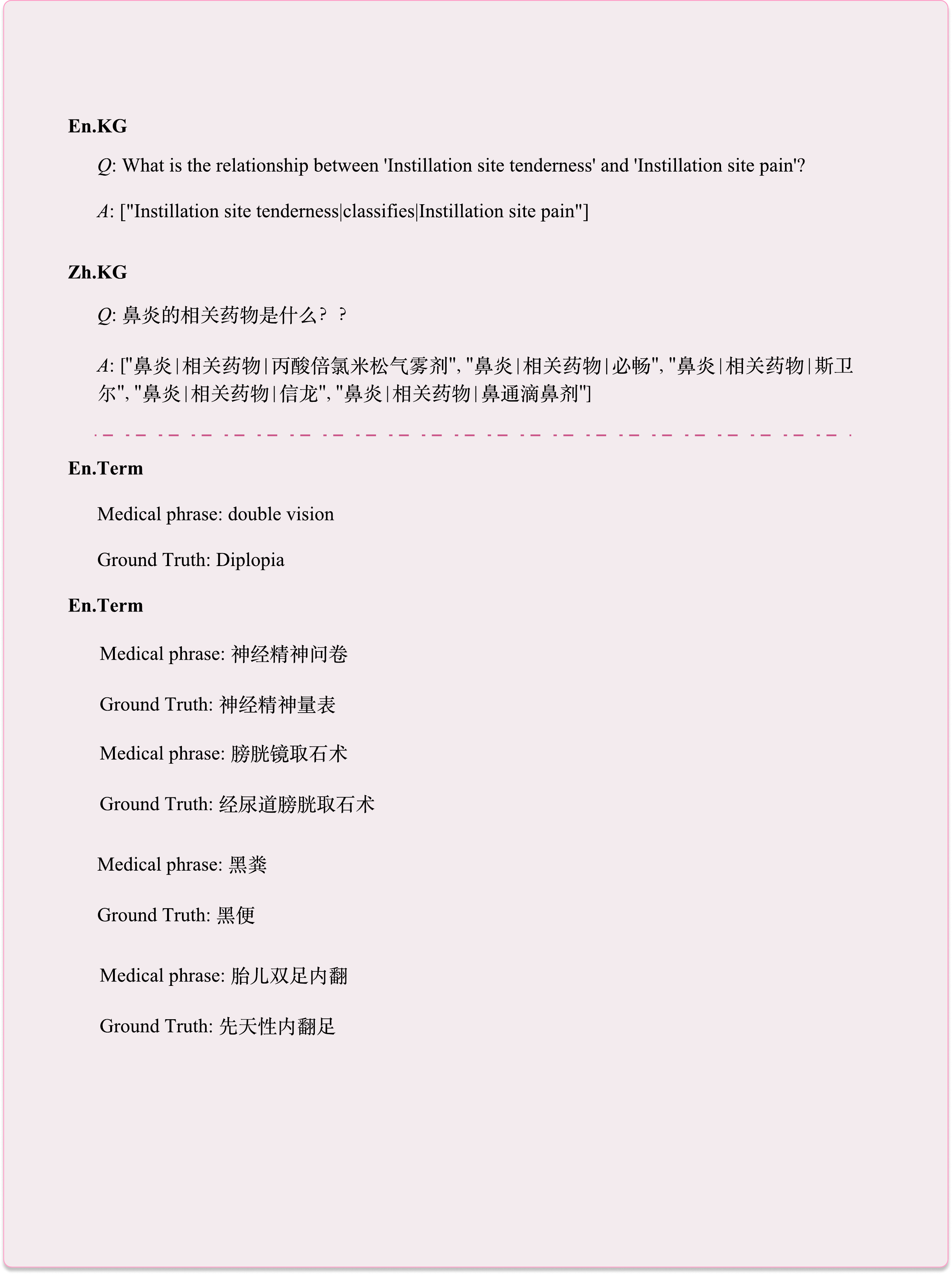}
    \caption{Examples of KG QA and Terminology Normalization.}
    \label{fig:kg_term_example}
\end{figure}

\begin{figure}[ht]
    \centering
    \includegraphics[width=\linewidth]{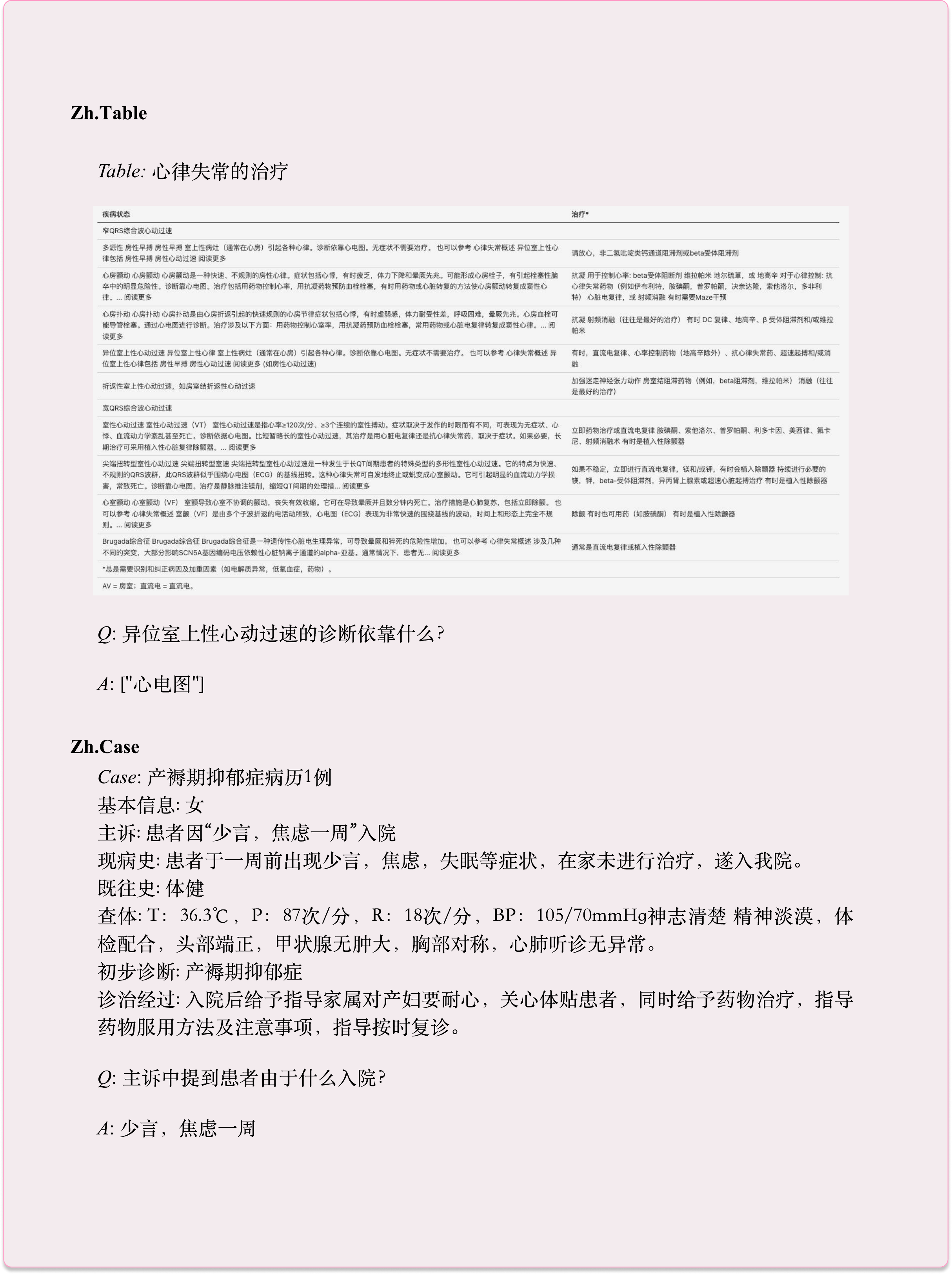}
    \caption{Examples of Table QA and Case QA.}
    \label{fig:table_case_example}
\end{figure}

\begin{figure}[ht]
    \centering
    \includegraphics[width=\linewidth]{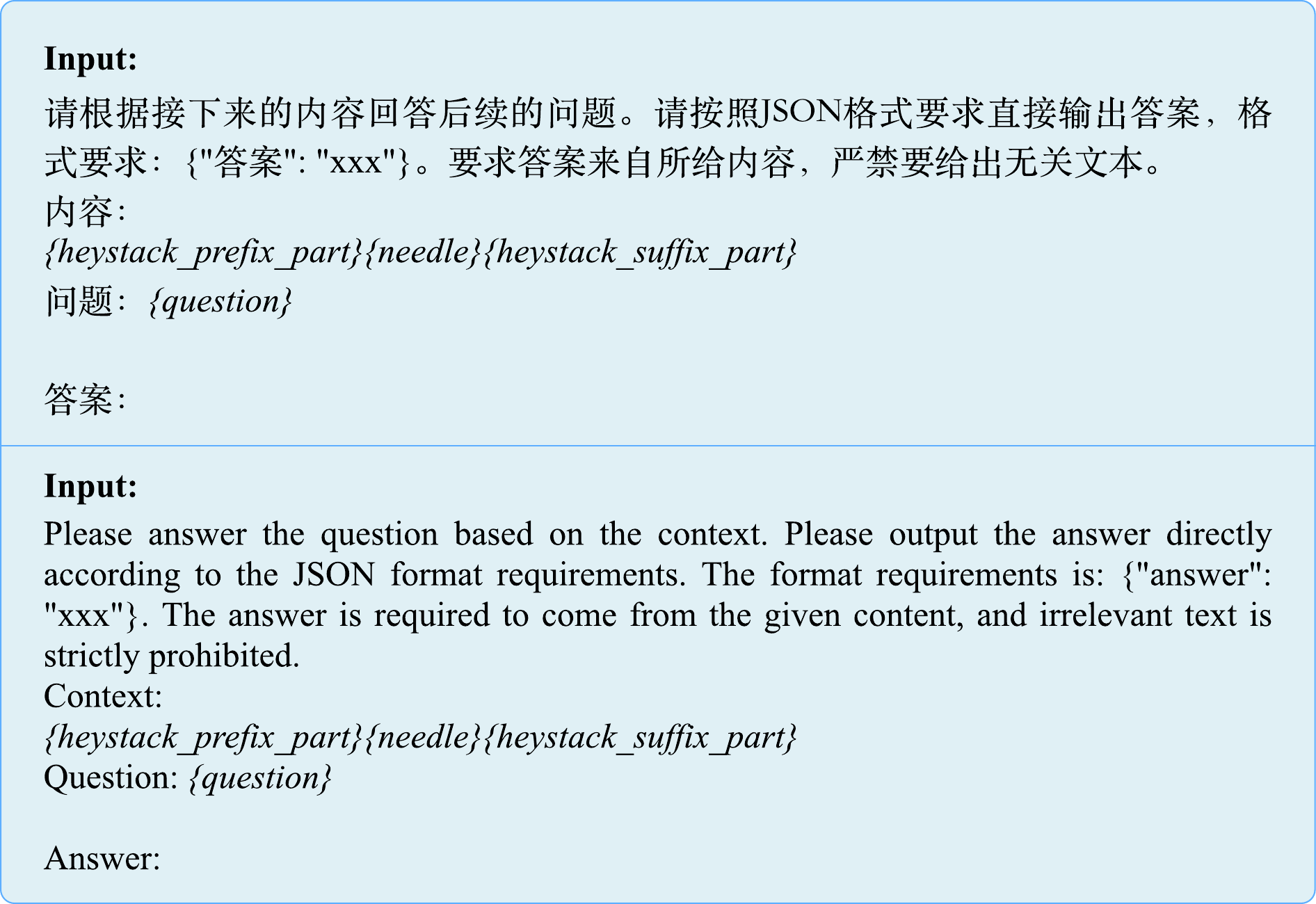}
    \caption{Prompt of the NIAH Tasks.}
    \label{fig:task_niah_prompt_example}
\end{figure}

\begin{figure}[ht]
    \centering
    \includegraphics[width=\linewidth]{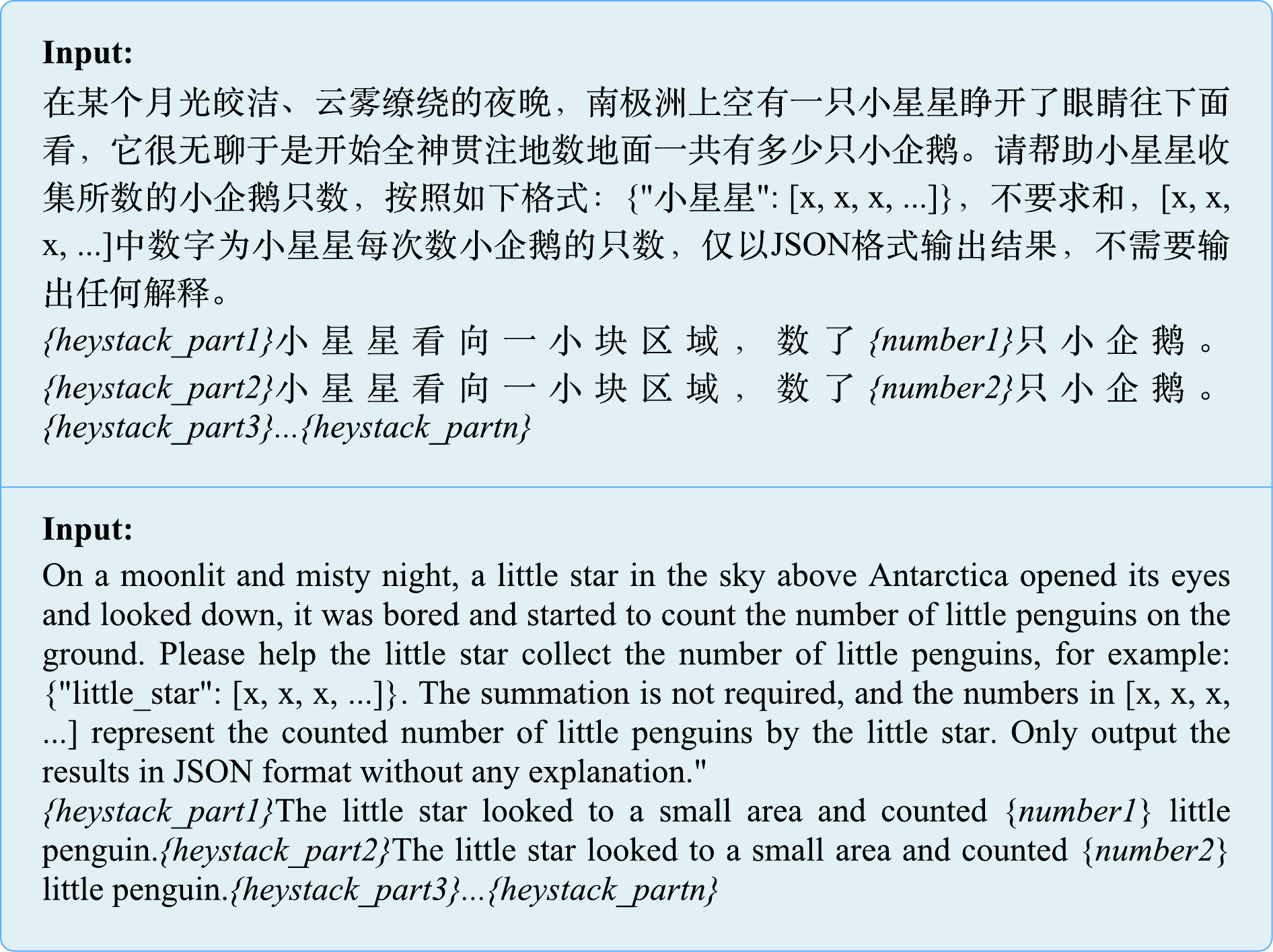}
    \caption{Prompt of the Counting Tasks (Type of Rep., Inc., and Shuf.).}
    \label{fig:task_counting_acqu_prompt_example}
\end{figure}

\begin{figure}[ht]
    \centering
    \includegraphics[width=\linewidth]{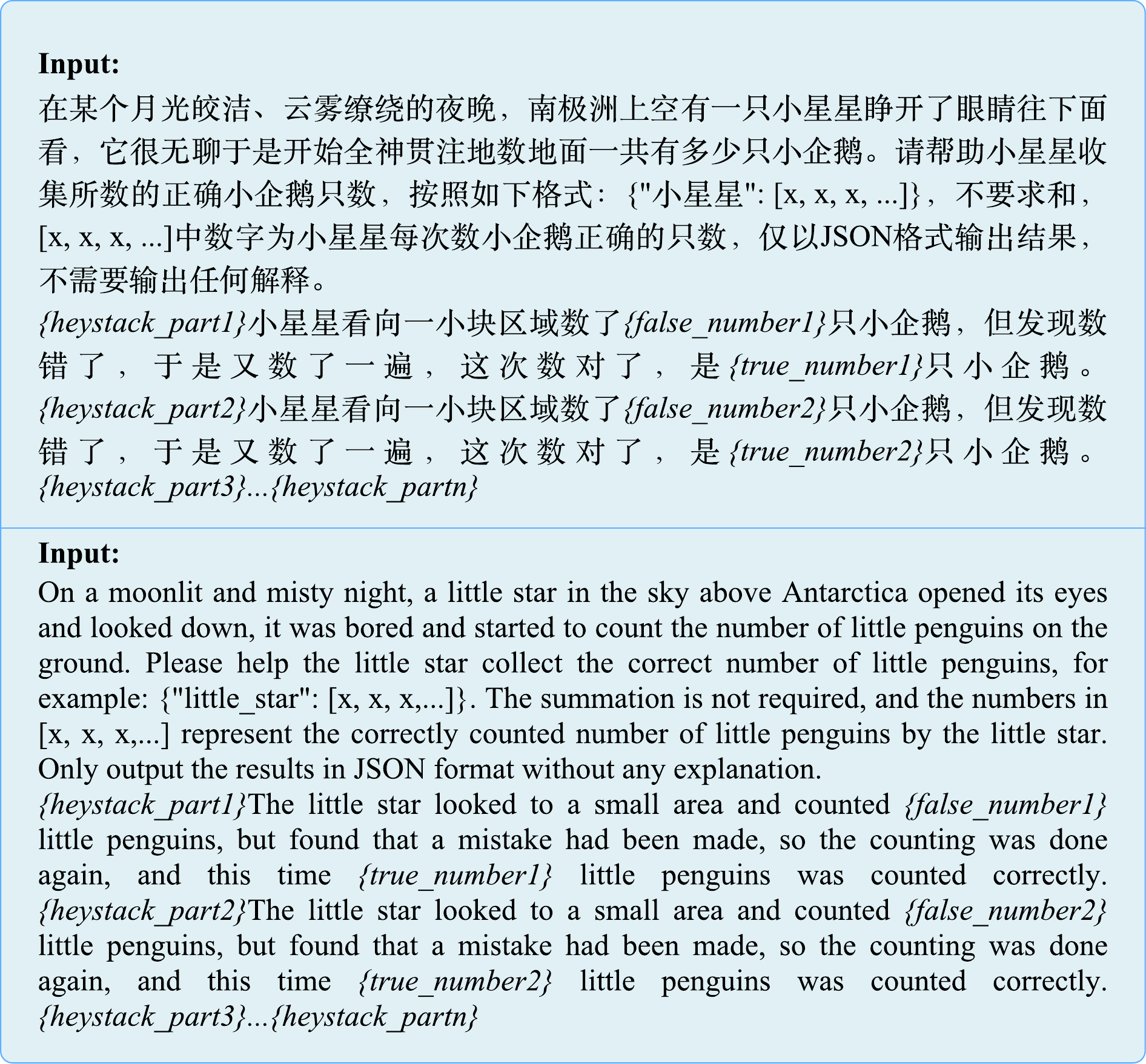}
    \caption{Prompt of the Counting Tasks (Type of Cor.).}
    \label{fig:task_counting_cof_prompt_example}
\end{figure}

\begin{figure}[ht]
    \centering
    \includegraphics[width=\linewidth]{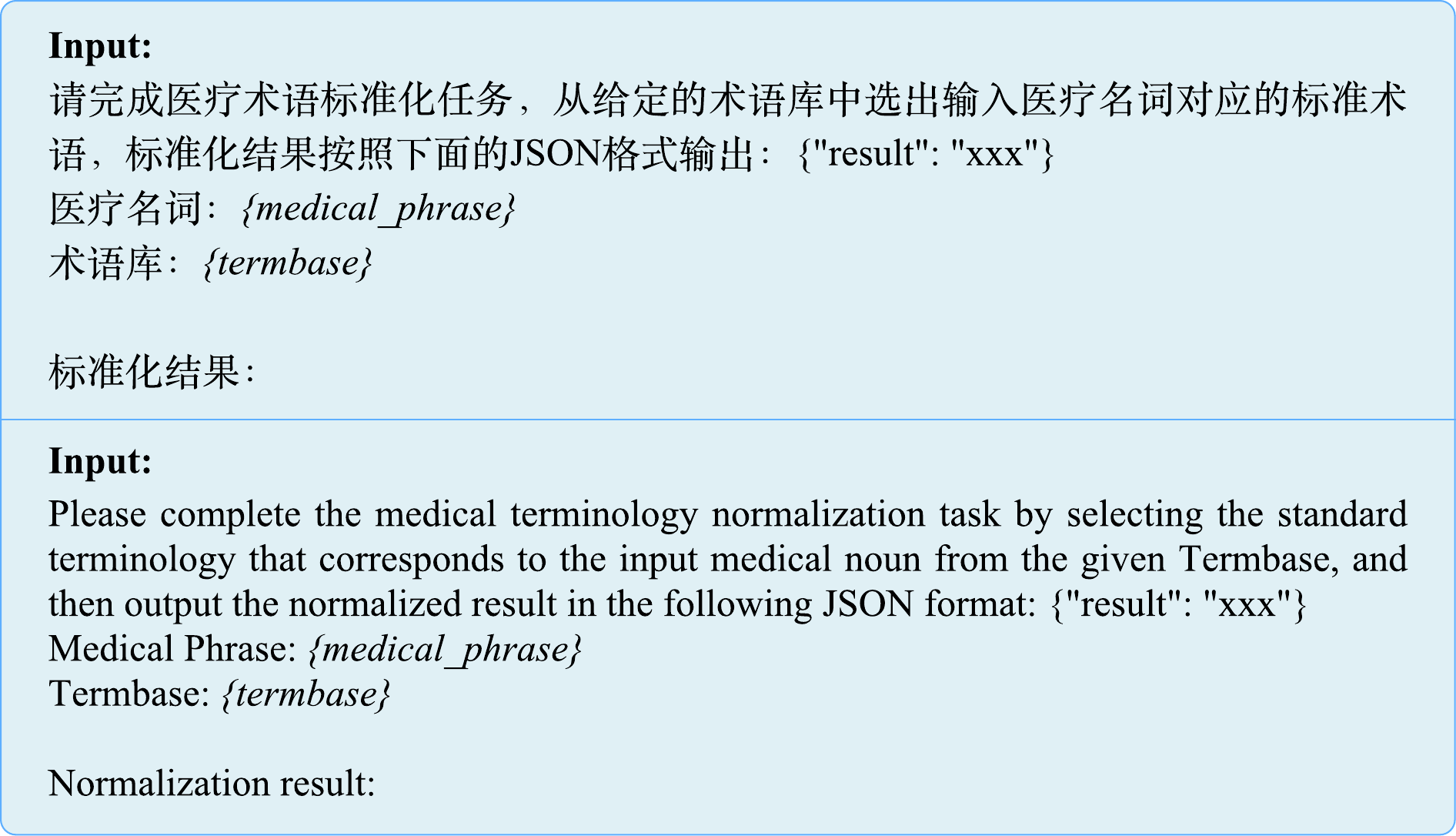}
    \caption{Prompt of the Term Tasks.}
    \label{fig:task_term_prompt_example}
\end{figure}

\begin{figure}[ht]
    \centering
    \includegraphics[width=\linewidth]{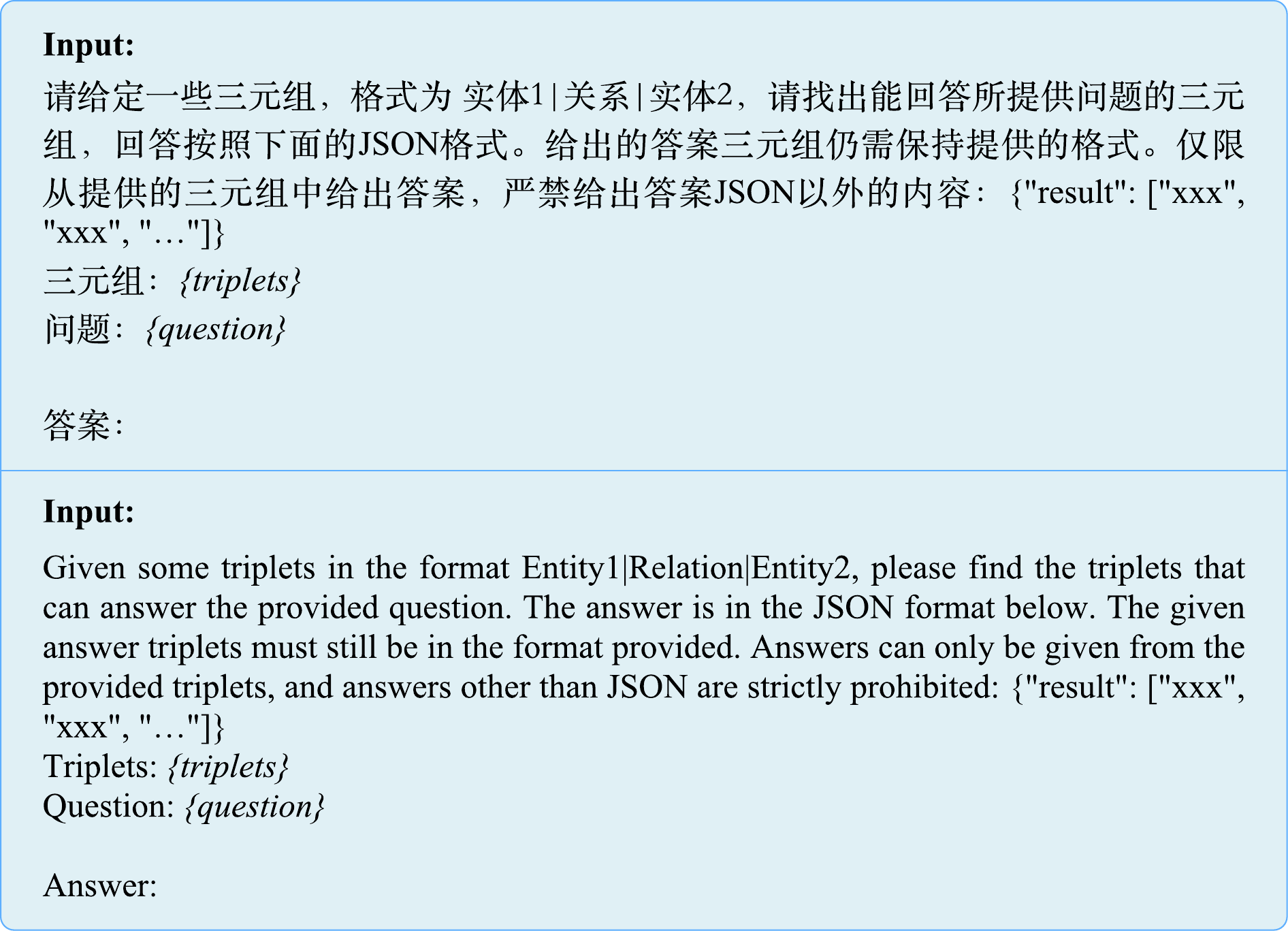}
    \caption{Prompt of the KG Tasks.}
    \label{fig:task_kg_prompt_example}
\end{figure}

\begin{figure}[ht]
    \centering
    \includegraphics[width=\linewidth]{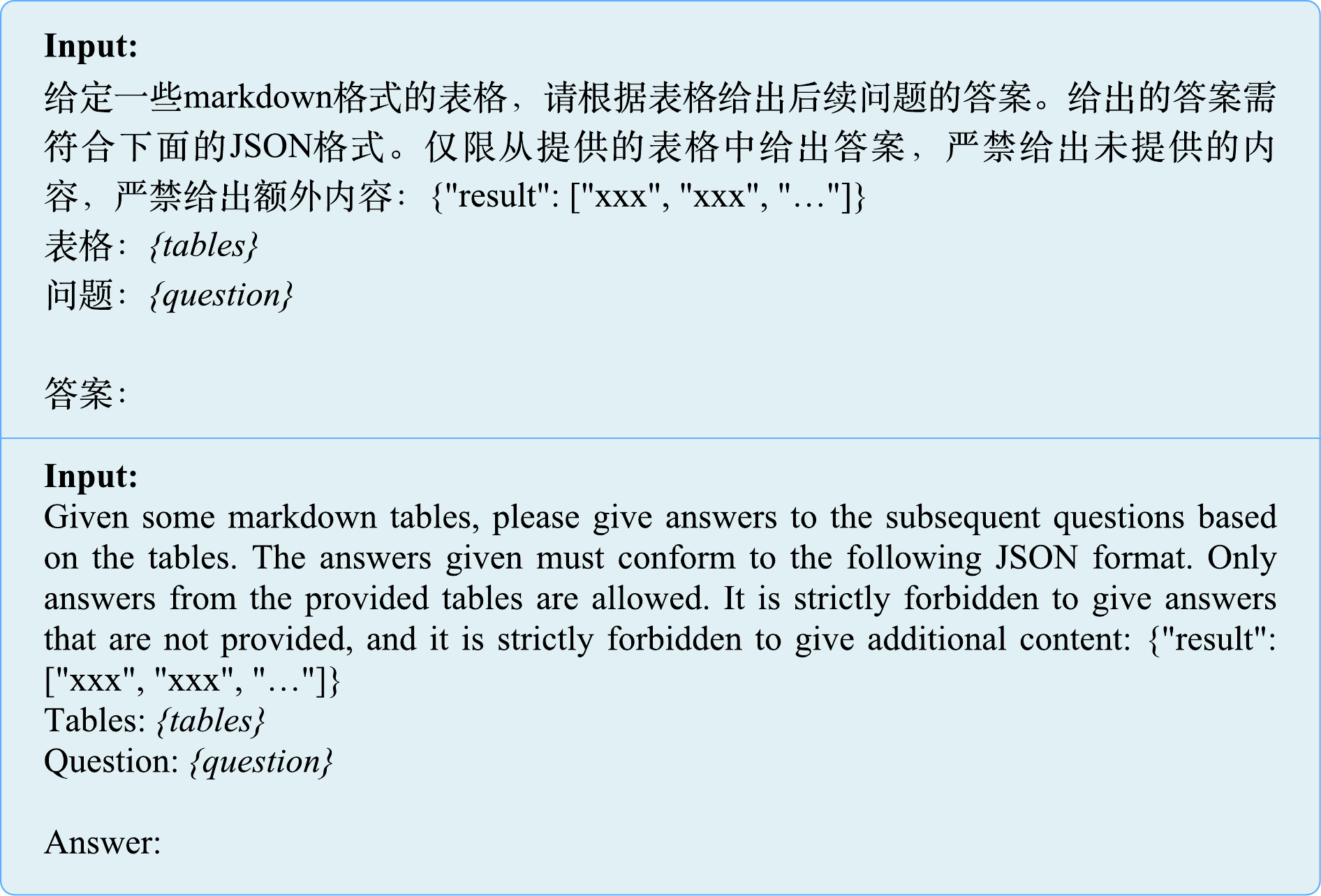}
    \caption{Prompt of the Table Tasks.}
    \label{fig:task_table_prompt_example}
\end{figure}

\begin{figure}[ht]
    \centering
    \includegraphics[width=\linewidth]{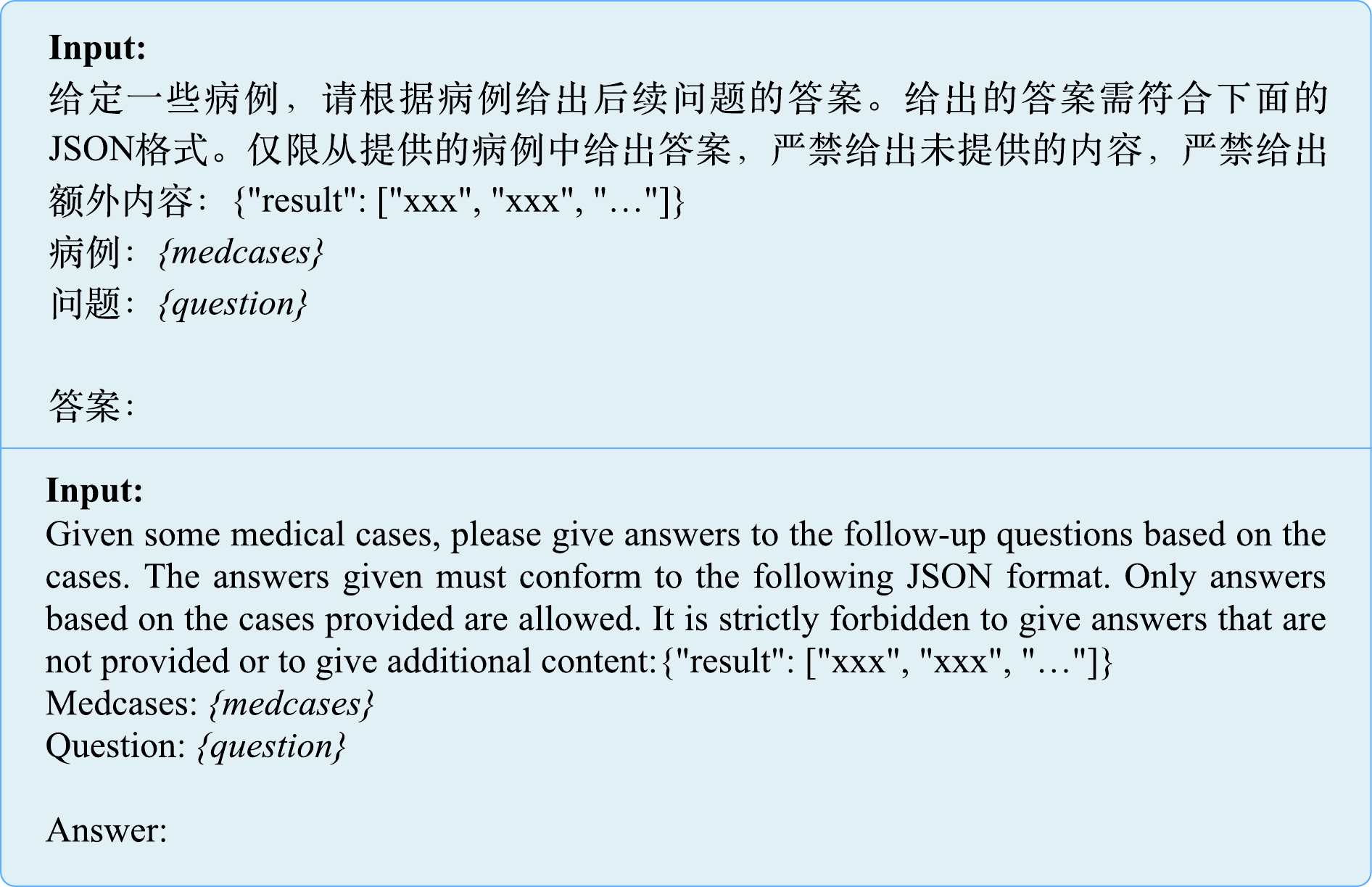}
    \caption{Prompt of the Case Tasks.}
    \label{fig:task_case_prompt_example}
\end{figure}

\end{document}